\documentclass[11pt,a4paper]{article}

\usepackage{amsmath,amssymb,amsthm,amscd,verbatim,enumerate}
\usepackage{graphicx,subfigure}
\usepackage[utf8]{inputenc}
\usepackage[colorlinks=true,citecolor=black,urlcolor=blue]{hyperref}
\usepackage[lmargin=31mm,rmargin=31mm,bmargin=31mm,tmargin=31mm]{geometry}
\usepackage{multirow}
\usepackage{xspace}
\usepackage{cancel}
\usepackage[small, width=\columnwidth]{caption}

\makeatletter
\let\@fnsymbol\@alph
\makeatother

\allowdisplaybreaks

\begin{document}

\title{Estimation of Human Body Shape and Posture Under Clothing}

\author{Stefanie Wuhrer\thanks{Cluster of Excellence MMCI, Saarland University, Germany, swuhrer@mmci.uni-saarland.de} \and Leonid Pishchulin\thanks{Max-Planck Institute for Informatics, Germany, leonid@mpi-inf.mpg.de} \and Alan Brunton\thanks{Fraunhofer Institute for Computer Graphics Research, Germany, alan.brunton@igd.fraunhofer.de} \and Chang Shu\thanks{National Research Council of Canada, chang.shu@nrc-cnrc.gc.ca} \and Jochen Lang\thanks{School of Electrical Engineering and Computer Science, University of Ottawa, Canada, jlang@eecs.uottawa.ca}}

\date{}

\maketitle

\abstract{
Estimating the body shape and posture of a dressed human subject in motion represented as a sequence of (possibly incomplete) 3D meshes is important for virtual change rooms and security. To solve this problem, statistical shape spaces encoding human body shape and posture variations are commonly used to constrain the search space for the shape estimate. In this work, we propose a novel method that uses a posture-invariant shape space to model body shape variation combined with a skeleton-based deformation to model posture variation. Our method can estimate the body shape and posture of both static scans and motion sequences of human body scans with clothing that fits relatively closely to the body. In case of motion sequences, our method takes advantage of motion cues to solve for a single body shape estimate along with a sequence of posture estimates. We apply our approach to both static scans and motion sequences and demonstrate that using our method, higher fitting accuracy is achieved than when using a variant of the popular SCAPE model~\cite{anguelov_srinivasan_koller_thrun_rodgers_05_shapecomp,Jain_etal_10_movie_reshape} as statistical model.
}


\section{Introduction}

The problem of estimating the body shape and posture of a dressed human subject is important for various applications, such as virtual change rooms and security. For instance, in virtual change rooms, a dressed user steps in front of a virtual mirror and the system aims to simulate different types of clothing for this user. To this end, such a system requires an accurate estimate of the body shape and posture of the user.

We present an algorithm to estimate the human body shape and posture under clothing from single or multiple 3D input frames that are corrupted by noise and missing data. Our approach assumes that the clothing fits to the body and may fail for loose clothing, such as skirts or wide dresses. When multiple 3D frames of the same human subject are recorded in different postures, these observations provide important cues about the body shape of the subject. The clothing may be more or less loosely draped around a particular body part in different postures, which allows for improved shape estimates based on postures where the clothing is close to the body shape. To utilize these cues, we model body shape independently of body posture, and optimize a single representation of the body shape of the subject along with one pose estimate per frame to fit to a set of input frames. When multiple 3D frames of a subject in motion are recorded with high frame rates, our algorithm takes advantage of the temporal consistency of the acquired data. To reduce the complexity of the problem, our method does not explicitly simulate the clothing, but learns information about likely body shapes using machine learning.

Current solutions to this problem use the SCAPE model~\cite{anguelov_srinivasan_koller_thrun_rodgers_05_shapecomp} to represent the body shape and posture of a human subject. This model represents the body shape in a statistical shape space learned from body scans of multiple subjects acquired in a standard posture and combines this with a representation of body posture learned from body scans of a single subject in multiple postures. A popular variant of SCAPE that performs well in practice is the method by Jain et al.~\cite{Jain_etal_10_movie_reshape} that learns variations in body posture using a skeleton-based deformation. The main disadvantage of these methods is that even when acquiring multiple subjects in standard posture, the postures differ slightly, which leads to a statistical space for body shape that represents a combination of shape and posture changes. Hence, for SCAPE and its variants, shape and posture representations are not properly separated.

To remedy this problem, we propose a method that uses a posture-invariant statistical shape space to model body shape combined with a skeleton-based deformation to model body posture. Using a posture-invariant statistical shape space for body shape offers the additional advantage that the shape space can be learned based on body scans of multiple subjects acquired in multiple postures, thereby allowing to leverage more of the available training data. 

This work makes the following main contributions:
\begin{itemize}
\item We present a representation that models human body shape and posture independently. Human body shape is represented by a point in a posture-invariant shape space found using machine learning, and human body posture is represented using skeletal joint angles.
\item We present an algorithm to estimate body shape and posture under clothing that fits closely to the body from single or multiple 3D input frames. For multiple input frames, a single representation of body shape is optimized along with a posture estimate per frame to fit to the input frames. This allows to take advantage of important cues about body shape from multiple frames.
\item When multiple 3D frames of a subject in motion are recorded with high frame rates, the presented fitting approach is stable as temporal consistency is used for tracking.
\item We show experimentally that using our method, higher fitting accuracy is achieved than when using the state of the art variant of SCAPE by Jain et al.~\cite{Jain_etal_10_movie_reshape}.
\end{itemize}

\section{Related Work}

The problem of estimating the body shape and posture of humans occurs in many applications and has been researched extensively in computer vision and computer graphics. Many methods focus on estimating the posture of a subject in an image or a 3D scan \emph{without} aiming to predict the body shape (e.g.~\cite{Horaud_etal_09,2011_BaakMuBhSeTh_DataDrivenFullBodyPose_ICCV,2011_StollEtAl_ICCV}). Other methods aim to track a \emph{given} human shape that may include detailed clothing across a sequence of images or 3D scans in order to capture the acquired motion without using markers (e.g.~\cite{Corazza_etal_06,deAguiarCVPR2007,Vlasic:2008:AMA,aguiar_etal_track_siggraph_08,gall_motion_2009}). 

In this work, we are interested in estimating both the body shape and posture of \emph{any} human subject represented as a 3D mesh that was acquired while wearing clothing. To achieve this goal, we need a model that can represent different body shapes in different postures. Statistical shape models have been shown to be a suitable representation in this case.

Statistical shape models learn a probability distribution from a database of 3D shapes. To perform statistics on the shapes, the shapes need to be in full correspondence. Allen et al.~\cite{allen_curless_popovic_03_parametrization_body_shape} proposed a method to compute correspondences between human bodies in a standard posture and to learn a shape model using principal component analysis (PCA). This technique has the drawback that small variations in posture are not separated from shape variations. To remedy this, multiple follow-up methods have been proposed. Hasler et al.~\cite{HasStoSunRosSei09} analyze body shape and posture jointly by performing PCA on a rotation-invariant encoding of the model's triangles. While this method models different postures, it cannot directly be constrained to have a constant body shape and different poses for the same subject captured in multiple postures. With the goal of analyzing body shape independently of posture, Wuhrer et al.~\cite{wuhrer_shu_xi_12_smi} propose to perform PCA on a shape representation based on localized Laplace coordinates of the mesh. In this work, we combine this shape space with a skeleton-based deformation model that allows to vary the body posture.

Several methods have been proposed to decorrelate the variations due to body shape and posture changes, which allow to vary body shape and posture independently. The most popular of these models is the SCAPE model~\cite{anguelov_srinivasan_koller_thrun_rodgers_05_shapecomp}, which combines a body shape model computed by performing PCA on a population of 3D models captured in a standard posture with a posture model computed by analyzing near-rigid body parts (corresponding to bones) of a single body shape in multiple postures. Chen et al.~\cite{Chen2013} recently proposed to improve this model by adding multi-linear shape models for each part of the SCAPE model, thereby enabling more realistic deformation behaviour near joints of the body. Neophytou and Hilton~\cite{Neophytou2013} proposed an alternative statistical model that consists of a shape space learned as PCA space on normalized postures and a pose space that is learned from different subjects in different postures. 

Several authors have proposed to use statistical shape models to estimate human body shape and posture under clothing. Most of these methods use the SCAPE model as statistical model. Muendermann et al.~\cite{MundermannCA07} proposed a method to track human motion captured using a set of synchronized video streams. The approach samples the human body shape space learned using SCAPE and initializes the body shape of the subject in the video to its closest sample in terms of height and volume. The approach then tracks the pose of the subject using an iterative closest point method, where joints are modeled as soft constraints. Balan and Black~\cite{BalanB08} used the SCAPE model to estimate the body shape and posture of a dressed subject from a set of input images. The method proceeds by optimizing the shape and posture parameters of the SCAPE model to find a human body that optimally projects to the observed silhouettes. If the same subject is given in multiple poses, the shape of the subject is assumed to be constant across all poses, and the model optimizes one set of shape parameters and several sets of posture parameters to fit the model to the observed input images. Weiss et al.~\cite{Weiss2011} used a similar technique to fit a SCAPE model to a Kinect scan. Zhou et al.~\cite{Zhou_etal_SIGGRAPH_10} used a SCAPE model to modify an input image. They learned a correlation between the SCAPE model parameters and semantic parameters, such as the body weight, which allows them to modify an instance of the SCAPE model to appear to have higher or lower body weight. The approach first optimizes a learned SCAPE model to fit to the input image, changes the shape of the 3D reconstruction of the subject, and modifies the input image, such that the silhouette of the modified subject is close to the projection of the changed 3D shape. Jain et al.~\cite{Jain_etal_10_movie_reshape} extended this approach to allow for the modification of video sequences. They used a slightly modified version of the SCAPE model that does not learn a subject-specific pose deformation of the triangles. Helten et al.~\cite{Helten2013} proposed a real-time full body tracker based on the Kinect. They first acquire the shape of a subject in a fixed posture using a Kinect, and then track the posture of the subject over time using the modified SCAPE model by Jain et al.~\cite{Jain_etal_10_movie_reshape} while fixing the shape parameters. 

A notable exception to using the SCAPE model is the approach by Hasler et al.~\cite{HasSto2009}, which uses a rotation-invariant shape space~\cite{HasStoSunRosSei09} to estimate body shapes under clothing. Recently, Perbet et al.~\cite{Perbet2014} proposed an approach based on localized manifold learning that was shown to lead to accurate body shape estimates. While these methods have been shown to perform well on static scans, they are less suitable to predict body shape and postures from motion sequences as the body shape cannot be controlled independently of posture in these shape spaces.

In this work, we are interested in fitting a single body shape estimate and multiple body posture estimates to a given sequence of scans, which requires a shape space that models variations of body shape and posture independently. The variant of the SCAPE model proposed by Jain et al.~\cite{Jain_etal_10_movie_reshape} is a commonly used state-of-the-art method that has been shown to lead to accurate body shape and posture estimates and that models shape and posture variations independently. We propose a new shape space that combines a posture-invariant statistical shape model with a skeleton-based deformation, and show that this model can fit more accurately to 3D input meshes than this popular variant of the SCAPE model.

\section{Overview}
\label{sec:overview}

We aim to estimate the body shape and postures of a dressed human in motion given as a set of $n$ input frames $F_1, \ldots, F_n$ represented as 3D points clouds. To solve this problem, our approach proceeds in two main steps. 

\paragraph{Training} We learn a statistical model based on a database of $k$ input scans denoted by $S_1, \ldots S_k$. To perform statistics on this database, all models of the database need to be in full point-to-point correspondence. While in general, computing correspondences between 3D models is a challenging problem~\cite{TamToappear}, template fitting approaches can be used in case of human models~\cite{allen_curless_popovic_03_parametrization_body_shape,HasStoSunRosSei09,posture_invariant_correspondence_11}. In this work, we use the registered publicly available MPI human shape database~\cite{HasStoSunRosSei09} (which contains a total of 520 models of over 100 subjects in up to 35 different postures) as training data. We learn two types of variations from the registered database. The first type of variation is information about a small set of landmark positions placed on the models, which helps in automatically detecting the corresponding landmarks on frames of a given motion sequence. These detected landmarks are then used to guide our model fitting. The second type of variation is a body shape model that captures body shape variations across different subjects in a posture-invariant way. This model has the advantage of capturing localized shape variations at the cost that it cannot be described using a small number of global linear mappings (such as SCAPE, for instance).

\paragraph{Fitting} We fit the learned statistical models to a given motion sequence $F_1, \ldots, F_n$. As the shape model cannot be described using a global linear mapping, we cannot directly fit this model to the data efficiently. To remedy this, the fitting procedure uses a rigged template $T$ with manually annotated landmarks and consists of four steps. First, we automatically predict landmark positions on the input frames $F_2, \ldots, F_n$ based on the learned space of landmark positions and given landmarks on the first frame $F_1$. Second, these landmark positions are used to consecutively fit the posture of $T$ to the postures of $F_i$ using a variational approach. Third, the shape of the template model is fitted to the input frames $F_i$ using a variational approach that allows the shape of $T$ to fit to details of clothing. After this fitting step, we have a sequence of deformed template shapes $T_1, \ldots, T_n$ that fit closely to the input frames $F_1, \ldots, F_n$. Note that $T_1, \ldots, T_n$ may not represent realistic body shapes, as the shapes may include geometric detail from the clothing. To remedy this, we restrict the shapes of $T_1, \ldots, T_n$ in a fourth step to a single point in the learned posture-invariant body shape space.

\section{Training a Posture-Invariant Statistical Model}

\begin{figure}[tb]
\centering
\includegraphics[width=12.0cm]{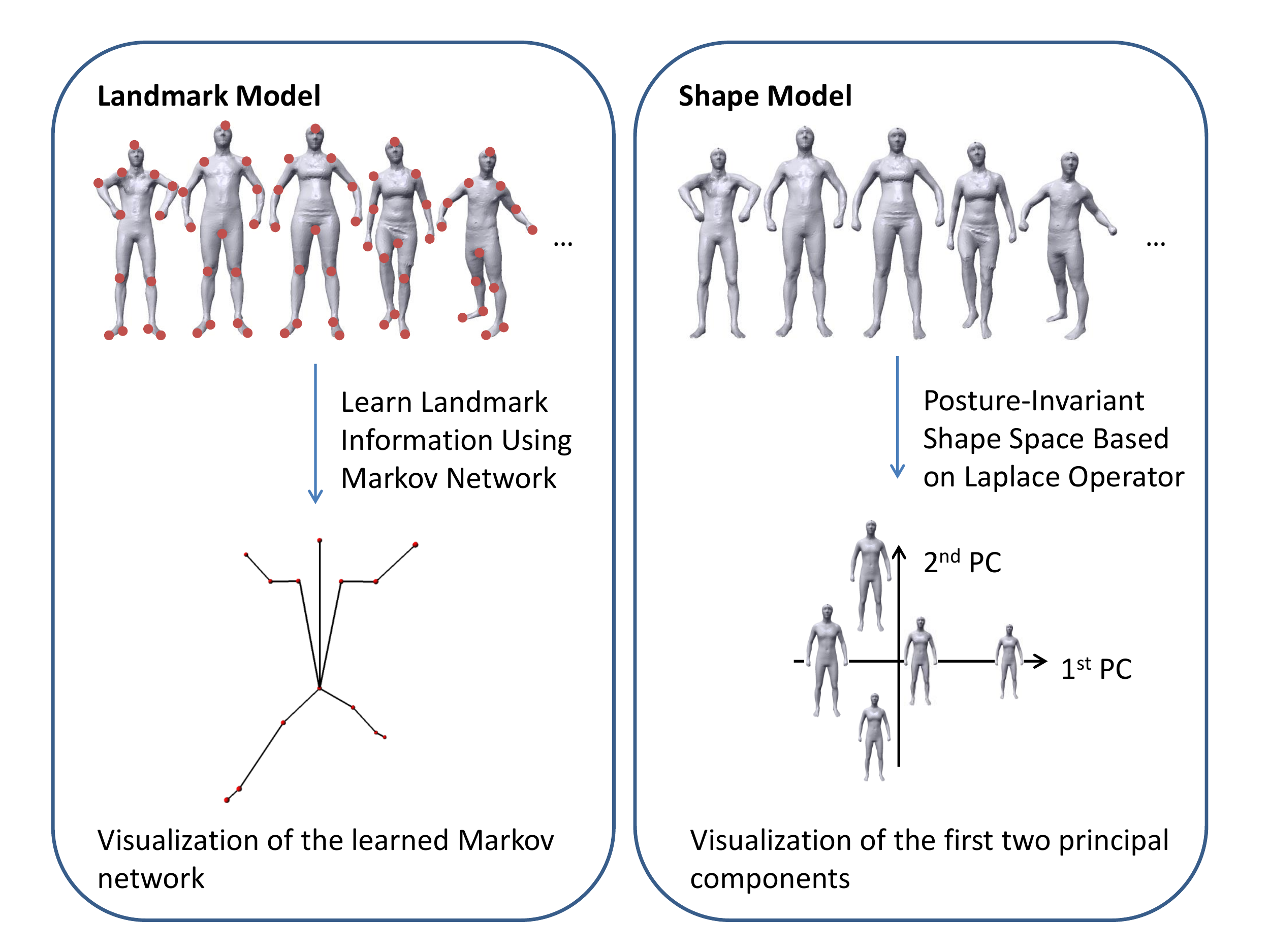}
\caption{Overview of training required by our method.}
\label{fig:training}
\end{figure}

This section outlines how to learn a statistical model based on a database of $k$ registered input scans $S_1, \ldots S_k$. Figure~\ref{fig:training} gives a visual overview of the two types of shape variations that are learned.

\subsection{Landmark Model}
\label{sec:train_MRF}

We use a Markov network to learn relative locations and local surface properties of the 14 anthropometric landmarks $l_1, \ldots, l_{14}$ shown as red points on the body shapes on the top left of Figure~\ref{fig:training}. We follow the approach of Wuhrer et al.~\cite{posture_invariant_correspondence_11}, which uses the network structure shown on the bottom left of Figure~\ref{fig:training}, where each red point represents a landmark, which is modeled as a node of the Markov network, and each black edge represents a connection between two landmark points, which is modeled as an edge of the Markov network. The approach uses a training database to learn the following node and edge potentials.

\paragraph{Node Potential} The approach learns a surface descriptor $d_{S_i,l_j}(a)$ for each landmark $l_j$ of input scan $S_i$ as the area of the geodesic neighborhood of radius $a$ centered at $l_j$ divided by the area of a planar disk of radius $a$. Note that $d_{S_i,l_j}(a)$ is invariant under isometric deformations, which are deformations that do not cause geometric stretching. Since the surface of a human body in different postures exhibits only limited and localized stretch, we can expect the descriptor $d_{S_i,l_j}(a)$ to be approximately posture-invariant. To learn localized surface properties around landmark $l_j$, $d_{S_i,l_j}(a)$ is computed for 20 radii $a_k$ from $1cm$ to $20cm$ over all input models $S_i$, and a multivariate Gaussian distribution is fitted to these descriptors. This multivariate Gaussian distribution is used as node potential for $l_j$ in the Markov network.

\paragraph{Edge Potential} The approach learns information about the spatial relationships between landmarks modeled as edge potentials. To learn this information, we first need to spatially align the training models $S_i$. However, it is difficult to spatially align models of human subjects due to the large posture variation. Hence, we compute an isometry-invariant canonical form~\cite{elad_kimmel_03_blending_signatures} of each of the models in the database. The canonical forms of all the models have a similar posture and can be spatially aligned using a rigid transformation computed using the known landmark positions. We can then learn the locations and relative positions of the landmarks in the space of canonical forms. We use this information to compute the edge potentials of the Markov network by computing the lengths and directions of each edge over all aligned models $S_i$, and by fitting a multivariate Gaussian distribution to this data.

Since all of the information contributing to the Markov network is isometry-invariant, this approach learns posture-invariant information about the landmark locations, which enables us to predict landmarks in arbitrary postures.

\subsection{Shape Model}
\label{sec:train_shapeModel}

To represent human body shape, we learn a posture-invariant statistical shape model based on localized Laplace coordinates, as proposed by Wuhrer et al.~\cite{wuhrer_shu_xi_12_smi}. This model, which we summarize in the following, is learned by performing PCA of a population of human shapes in arbitrary postures using a posture-invariant shape representation, and visualized on the right of Figure~\ref{fig:training}.

This shape representation stores for each vertex of $S_i$ the Laplace offset in a local coordinate system. That is, we find a posture-invariant representation of $S_i$ by computing the combinatorial Laplace matrix $L$ of $S_i$. With the Laplace matrix, we can compute the Laplace offsets $\Delta_j$ as
\begin{equation}
\left( \begin{array}{c} \Delta_1\\ \ldots \\ \Delta_m\end{array} \right) = L \left( \begin{array}{c} v_1\\ \ldots \\ v_m\end{array} \right),
\end{equation}
where $v_1, \ldots, v_m$ denote the vertices of $S_i$. These offsets are not posture-invariant. Hence, we express each offset with respect to the following local coordinate system. At each vertex $v_j$, we pick an arbitrary but fixed neighbor $v_k$ as the first neighbor (we choose the same first neighbor for all of the parameterized meshes). We then compute a local orthonormal coordinate system at $v_j$ using the normal vector at $v_j$, the normalized projection of the difference vector $v_k-v_j$ to the tangent plane of $v_j$, and the cross product of the previous two vectors. We denote the three vectors defining the local orthonormal coordinate system by $f_1\left(v_j\right), f_2\left(v_j\right)$, and $f_3\left(v_j\right)$. Since the local coordinate system is orthonormal, we can express $\Delta_j$ in this coordinate system as 
\begin{equation}
\Delta_j = \omega_j^{1} f_1\left(v_j\right) + \omega_j^{2} f_2\left(v_j\right) + \omega_j^{3} f_3\left(v_j\right).
\end{equation} 
The local coordinates $\omega_j^{k}$ are designed to be invariant with respect to rigid transformations of the one-ring neighborhood of $v_j$. To account for global scaling of the shape, we also store a coefficient $s_i$ related to the scale of the shape. More specifically, $s_i$ is computed as the average geodesic distance between any two vertices on $S_i$ computed using the fast marching technique~\cite{kimmel_sethian_98_computing_geodesic}.

We then perform statistical shape analysis by performing PCA on the vectors $\left[\omega_j^{k}, s_i\right]^T$ over all shapes. Let $\mathcal{S}$ denote the learned posture-invariant shape space. To avoid problems related to over-fitting a statistical model, in this work, we keep only about $70\%$ of the shape variability present in the training set.

\section{Estimating Body Shape and Posture from Motion Sequences of Dressed Subjects}

\begin{figure}[tb]
\centering
\includegraphics[width=12.0cm]{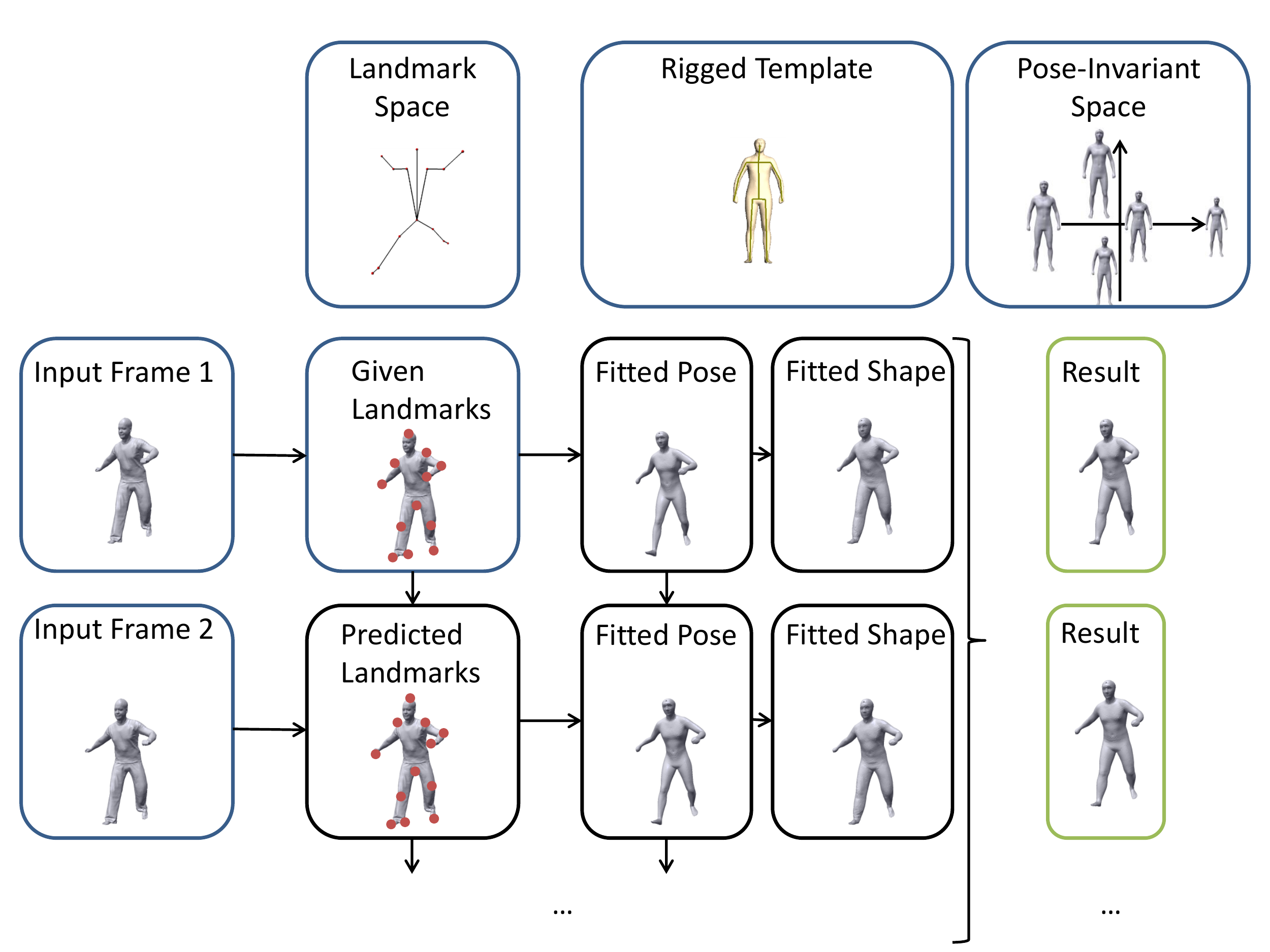}
\caption{Overview of fitting procedure. Blue boxes show the input to our method and green boxes show the results.}
\label{fig:fitting}
\end{figure}

This section describes our proposed approach to estimate the body shape and posture of a sequence of input meshes $F_1, \ldots, F_n$ showing a dressed human in motion. Ideally, we would like to fit the learned shape model to the data directly. However, this is not efficient because the posture-invariant shape model cannot be described using a small number of global linear transformations. Hence, we use a fitting procedure consisting of four steps. Figure~\ref{fig:fitting} gives a visual overview of the four steps of the approach. First, we use the learned Markov network to predict the locations of the 14 landmarks $l_j$. Specifically, we require the user to provide the locations of $l_j$ for $F_1$, and then predict $l_j$ on the remaining input frames $F_i$ automatically. The advantage of user-specified landmarks on the first frame is that the landmark tracking starts with a good initialization. Second, we use the landmark locations $l_j$ to fit the posture of the rigged template $T$ to the frames $F_i$. This deforms the skeleton of $T$ using a piecewise rigid transformation that is blended onto the surface of $T$. That is, $T$ is deformed using an approximately piecewise rigid transformation in this step. Third, we fit the body shape of $T$ to the observed data $F_i$ using a non-rigid deformation model, which allows for $T$ to deform closely to $F_i$. Let $T_i$ denote the deformation of $T$ that was fitted to $F_i$. Once the posture and shape of $T$ has been fitted to each of the input frames $F_i$, the resulting shapes $T_i$ may not represent realistic human body shapes because parts of $T_i$ may be close to data acquired from clothing. Fourth, to find a single realistic body shape estimate in multiple postures, we restrict the shapes of $T_i$ to a single point in the learned posture-invariant shape space. Figure~\ref{fig:fitting} shows results for each of the four steps for two input frames.

\subsection{Landmark Prediction}

We now outline how the landmark locations are predicted using probabilistic inference on the Markov network with learned potentials that is described in Section~\ref{sec:train_MRF}. Given an input mesh $F_i$, we need a set of possible labels, which represent possible locations for the landmark locations $l_j$ in order to perform probabilistic inference. For a possible label $l$ for location $l_j$, we can compute the node potential as $d_{F_i,l}(a_k)$ for the 20 possible values for the radii $a_k$ used for training, which allows to compute the probability of $l$ being the location of landmark $l_j$ on $F_i$. Given pairs of possible labels of landmarks that are connected by an edge in the Markov network, we can compute the edge potential by computing the distance between the two labels in the canonical form of $F_i$, which allows to compute the joint probability of the two labels being the locations of the corresponding landmarks. Since the graph representing the connections between the landmark locations is a tree, a simple message passing scheme can then be used to find the labels that maximize the joint probability of being the landmark locations~\cite[Chapter 4]{pearl_88}.

It remains to discuss how the sets of possible labels for landmark $l_j$ are found. Recall that we assume that the landmark locations on the first frame $F_1$ are provided by the user (this is the only user input assumed by our fitting algorithm). For the remaining frames, we take advantage of the temporal consistency of the input sequence to find sets of possible labels for $F_i$ based on the predicted landmark locations on frame $F_{i-1}$. That is, vertices on $F_i$ in the neighborhood of the predicted landmark $l_j$ on $F_{i-1}$ are considered as candidate labels for $l_j$. In our implementation, we choose as label set the 200 points on on $F_i$ that are closest to $l_j$. 

This selection of the label set, which is the main difference to the landmark prediction method by Wuhrer et al.~\cite{posture_invariant_correspondence_11} that predicts landmarks on a static scan $F$ using label sets found using the canonical form of $F$, has two advantages. First, our approach is computationally more efficient than the previous method as, thanks to the temporal consistency between adjacent frames, a single label set suffices to predict landmarks accurately. In contrast, the method by Wuhrer et al. considers eight label sets found using eight possible alignments in canonical form space, computes a candidate solution for each label set using probabilistic inference, and finally selects the most suitable solution automatically using an energy term. Second, our approach is designed to lead to stable solutions as corresponding landmarks in adjacent frames are close to each other, which prevents prediction errors due to symmetric regions (i.e. mixing up the left and right sides of the body).

Hence, by design, the tracking of the landmarks is robust with respect to changes that have the property that each landmark on $F_i$ is in the neighborhood of its corresponding landmark on $F_{i-1}$. We validate experimentally that this assumption holds for human motion sequences even in the presence of fast localized movements. Note that since we perform probabilistic inference on the learned Markov network to find the best landmark location, the landmark on $F_i$ does not need to be the closest neighbor to its corresponding landmark on $F_{i-1}$.

\subsection{Posture Fitting}

Given a set of (predicted) landmarks $l_j$ on $F_i$, we aim to fit the posture of a rigged template model $T$ to the posture of $F_i$. We compute our template $T$ as the mean shape over all models of the training database that were captured in a standard posture. The model $T$ is rigged using the publicly available software Pinocchio~\cite{baran_popovic_07_animation}, and the landmark locations $l_j$ are manually placed on $T$.

We model the deformation of the skeleton of $T$ using a scene graph structure consisting of 17 bones, where bones are ordered in depth first order, and the transformation of each bone is expressed using a local transformation relative to its parent. The bone structure of the rigged template is shown in the top row of Figure~\ref{fig:fitting}. The root bone is transformed using a rigid transformation consisting of a rotation (parameterized using a rotation axis and angle), a scale factor, and a translation vector. The relative transformations of the remaining bones are expressed using a rotation with respect to their parent bones. We denote the transformation parameters of the bones by $b_k$. Note that it is straight forward to compute the global bone transformations $\textbf{B}_k$ using composite transformations.

Our posture fitting method extends the variational approach proposed by Wuhrer et al.~\cite{posture_invariant_correspondence_11}, which estimates the posture of a static scan, to estimate a sequence of postures for a given set of frames. To find posture estimates that are stable over time efficiently, we take advantage of the temporal consistency between adjacent frames. That is, we initialize the transformation parameters $b_k$ of frame $F_i$ to the final result computed for frame $F_{i-1}$ for $i>1$. This initialization not only ensures that the resulting posture estimates change smoothly over time, but also leads to an efficient optimization as the initial posture parameters are generally close to the optimal solution. We validate experimentally that this initialization allows to accuractely estimate the postures even in the presence of fast localized movements.

With this initialization, we proceed as in the static case by optimizing the posture using two consecutive energy minimizations. First, we use the anthropometric landmark locations to optimize the posture by minimizing
\begin{equation}
E_{lnd} = \sum_{j=1}^{14} \left\| \left( \sum_{k=1}^{17} w_{j,k}\textbf{B}_k l_j^{(T)} \right) - l_j^{(F_i)}\right\|^2
\end{equation}
with respect to the parameters $b_k$, where $w_{j,k}$ is the rigging weight for the $k$-th bone and the $j$-th landmark of $T$, $l_j^{(T)}$ denotes landmark $j$ on $T$, and $l_j^{(F_i)}$ denotes landmark $j$ on the current frame $F_i$. 

Second, we use all vertex positions on frame $F_i$ to optimize the posture by minimizing
\begin{equation}
E_{nn} = \sum_{j} \left\| \left( \sum_{k=1}^{17} w_{j,k}\textbf{B}_k v_j \right) - NN(v_j) \right\|^2
\end{equation}
with respect to the parameters $b_k$, where $w_{j,k}$ is the rigging weight for the $k$-th bone and the $j$-th vertex of $T$ and where $NN(v_j)$ is the nearest neighbor of the transformed vertex $\left( \sum_{j=1}^{17} w_{j,k}\textbf{B}_k v_j \right)$ in frame $F_i$. 

\subsection{Shape Fitting}

This section describes how to change the shape details of the posture-aligned template model to fit to the shape of frame $F_i$. To simplify notation, in this section, let $T$ denote the template model after it was deformed to match the posture of $F_i$.

The remaining problem is to fit $T$ to a frame $F_i$, where $F_i$ has a similar posture as $T$. We solve this problem using an energy optimization method similar to the one by Allen et al.~\cite{allen_curless_popovic_03_parametrization_body_shape}, who deform each vertex $v_j$ of $T$ using an affine transformation matrix $\textbf{A}_j$. That is, the deformed vertex is expressed as $v_j^* = \textbf{A}_j v_j$, and the goal is to find $\textbf{A}_j$ that moves every vertex of $T$ close to the scan $F_i$ while maintaining a smooth deformation field. The smoothness is modeled using the energy $\sum_{(j,k)\in E} \|\textbf{A}_j - \textbf{A}_k\|_F^2,$ where $E$ is the edge set of $T$ and where $\|.\|_F$ denotes the Frobenius norm. 

One drawback of this approach is that the Frobenius norm between transformation matrices is used to measure the difference between transformations. This is problematic because a global scaling of the object results in a different relative weighting of the rotation and translation components encoded in $\textbf{A}_j$.

We remedy this problem by deforming each vertex using a translation and a rotation. The translation is encoded using a translation vector $t_j$, and the rotation is encoded using a rotation axis $r_j$ and a rotation angle $\alpha_j$. Let $\textbf{A}(t_j)$ be the ($4\times 4$) matrix that translates a point by translation vector $t_j$, and let $\textbf{A}(r_j, \alpha_j)$ be the ($4\times 4$) matrix that rotates a point by angle $\alpha_j$ around $r_j$. We compute the deformation matrix $\textbf{A}_j$ as $\textbf{A}_j = \textbf{A}(v_j) \textbf{A}(r_j, \alpha_j) \textbf{A}(t_j) \textbf{A}(-v_j)$. That is, the deformation parameters are expressed with respect to a local coordinate frame centered at $v_j$. 

The goal is to fit $T$ to $F_i$ using a smooth deformation field by minimizing
\begin{eqnarray}
E_{shape} & = & \omega_{data} \sum_{j} \left\| \textbf{A}_j v_j - NN(v_j)\right\|^2 \nonumber \\
& + & \omega_{clothing} \sum_{j} \rho(v_j) \nonumber \\
& + & \omega_{smooth} \sum_{j} \sum_{k \in D_j} \left( 1-\frac{\|v_j - v_k\|^2}{d^2} \right) \nonumber \\
& \cdot & \left(\|t_j-t_k\|^2 + \left\| \frac{r_j}{\|r_j\|} - \frac{r_k}{\|r_k\|} \right\|^2 + (\alpha_j - \alpha_k)^2 \right) 
\label{eq:E_shape}
\end{eqnarray}
with respect to the deformation parameters $t_j, r_j$ and $\alpha_j$, where $NN(v_j)$is the nearest neighbor of the transformed vertex $\textbf{A}_j v_j$ in $F_i$, $d$ is twice the average edge length in $T$, and $D_j$ contains the set of all points of $T$ located within a sphere of radius $d$ centered at $v_j$. Here, $\rho(v_j)$ is a function that measures the distance of a vertex of the template to the interior of the frame $F_i$ as 
\begin{equation}
	\rho(v_j) = \begin{cases} n(NN(v_j))^T (\textbf{A}_j v_j - NN(v_j)) \mbox{ if } n(NN(v_j))^T (\textbf{A}_j v_j - NN(v_j)) > 0 \\ 0 \mbox{ otherwise,} \end{cases}
\end{equation}
where $n(NN(v_j))$ is the outer normal vector of point $NN(v_j)$ on $F_i$.

The first energy term drives the template mesh to the observed data. The second energy term encourages the template to stay within the volume of the observed scan $F_i$\footnote{We thank the anonymous reviewer for suggesting this energy term.}. A similar energy term has recently been introduced by Perbet et al.~\cite{Perbet2014}. We only consider the first two terms corresponding to $v_j$ if the angle between the outer normal vectors of the transformed vertex on the template and its nearest neighbor in the scan is at most $90$ degrees. The third energy term encourages a globally smooth deformation of the surface by encouraging close-by points (measured with respect to the local mesh resolution around the points) to have similar deformation parameters. For this energy term, points that are closer in the template mesh obtain a higher weight than points that are farther away.

We initialize $t_j$ to the zero vector, $r_j$ to the normalized vector pointing in direction $\left[1, 1, 1\right]^T$, and $\alpha_j$ to zero. Following previous work on template fitting~\cite{allen_curless_popovic_03_parametrization_body_shape,li08global}, our approach starts by setting $\omega_{data}$ and $\omega_{clothing}$ to a relatively low value compared to $\omega_{smooth}$ to smoothly deform $T$ towards $F_i$, and subsequently increases the relative influence of $\omega_{data}$ and $\omega_{clothing}$ to allow $T$ to fit more closely to $F_i$ in localized areas. Specifically, in our implementation, we initially set $\omega_{data} = 1$, $\omega_{clothing} = 1$, and $\omega_{smooth}^{(0)} = 5$, and we relax $\omega_{smooth}^{(t)}$ as $\omega_{smooth}^{(t)} = 0.5\omega_{smooth}^{(t-1)}$ whenever the energy does not change much. We stop if $\|E_{shape}^{(t-1)} - E_{shape}^{(t)}\| / E_{shape}^{(t-1)} < 0.001$ or $\omega_{smooth}^{(t)} < 0.1$.

\subsection{Restriction to Learned Shape Model}

After fitting $T$ to each frame $F_i$, we have a set of parameterized models. All of these models describe the same subject, and hence, they should all have the same body shape. However, if the subject we track was dressed during the acquisition, the shapes of some or all of the frames may include geometric detail that is not part of the human body shape. We now adjust the shapes such that they lie within the learned shape space of human body shapes. 

For simplicity, in the following let $T_i$ denote the parameterized frames found by minimizing Equation~\ref{eq:E_shape}. Using the learned posture-invariant shape space $\mathcal{S}$ from Section~\ref{sec:train_shapeModel}, we can express each $T_i$ as a point in $\mathcal{S}$. Recall that $\mathcal{S}$ was learned based on a set of training shapes $S_i$. If the tracking result found the accurate body shape for each frame, all $T_i$ should correspond to the same point in $\mathcal{S}$. However, in practice, due to the presence of noise and clothing, the points are different. We choose the mean of the projections of $T_i$ into $\mathcal{S}$ to represent the initial body shape estimate. Let $z$ denote this representative. If the user is willing to provide confidence weights for each frame that describe how closely the captured scan is to the true body shape, the representative $z$ can be computed as a weighted average, where each $T_i$ is weighted by the given corresponding confidence weight. Note that in general, $z$ is different from the mean of the learned PCA space $\mathcal{S}$. If $z$ is located far from the mean shape of the training population $S_i$ (which is the origin of $\mathcal{S}$), it is likely that clothing resulted in tracking results that do not accurately represent the body shape of the subject. In this case, we move $z$ to the intersection of the line through $z$ and the origin of $\mathcal{S}$ with the ellipsoid $x^T (3\Sigma)^{-1} x = 1$, where $\Sigma$ is the covariance matrix of the population $S_i$. That is, we move $z$ linearly towards the origin of $\mathcal{S}$ until $z$ is at most three standard deviations from the origin of $\mathcal{S}$.

The representative $z$ describes the body shape of the captured subject in $\mathcal{S}$. Using the learned principal components, we can compute the local coordinates $\omega_j^{k}$ and the scale $s$ corresponding to $z$. We now deform each frame $T_i$ to achieve these local coordinates and scale. Recall from Section~\ref{sec:train_shapeModel} that for any mesh
\begin{equation}
L \left( \begin{array}{c} v_1\\ \ldots \\ v_m\end{array} \right) = \left( \begin{array}{c} \omega_1^1 f_1(v_1) + \omega_1^2 f_2(v_1) + \omega_1^3 f_3(v_1) \\ \ldots \\ \omega_m^1 f_1(v_m) + \omega_m^2 f_2(v_m) + \omega_m^3 f_3(v_m)\end{array}\right).
\label{eq:reconstruct}
\end{equation}
Here, $L$, $\omega_i^j$ and $s$ are given and we aim to find vertex positions $v_j$ that satisfy the above equation. 

Equation~\ref{eq:reconstruct} implies that $v_j = \sum_{v_k \in \mathcal{N}_1(v_j)} \frac{1}{deg(v_j)} v_k - (\omega_j^1 f_1(v_j) + \omega_j^2 f_2(v_j) + \omega_j^3 f_3(v_j))$, where $N_1(v_j)$ is the one-ring neighborhood of $v_j$. Hence, we can find a solution by deforming the vertices $v_j$ of each frame $T_i$ to minimize 
\begin{equation}
E_{human} = \sum_{j} \left\|v_j - \sum_{v_k \in N_1(v_j)} \frac{1}{deg(v_j)} v_k - \left(\omega_j^1 f_1(v_j) + \omega_j^2 f_2(v_j) + \omega_j^3 f_3(v_j)\right)\right\|^2.
\label{eq:human}
\end{equation}

Wuhrer et al.~\cite{wuhrer_shu_xi_12_smi} optimize $E_{human}$ for a single frame using a two-stage process consisting of an iterative method followed by a quasi-Newton optimization that ensures that a good local minimum is found. In our case, however, the use of temporal consistency between adjacent frames during tracking results in frames $T_i$ that provide a good initialization for the quasi-Newton optimization of Equation~\ref{eq:human}. Hence, we can directly minimize $E_{human}$ using a quasi-Newton method, which leads to a gain in efficiency. 

\section{Evaluation}

We implemented the proposed approach using C++. To compute (exact) nearest neighbors, the implementation in ANN~\cite{arya_mount_93_ann} is used, and to minimize the energies $E_{lnd}$, $E_{nn}$, $E_{shape}$, and $E_{human}$, a quasi-Newton approach~\cite{liu_nocedal_lbfgsb} is used.

\subsection{Estimating Shape and Posture Using Static Scans}

We first evaluate our approach when fitting the proposed statistical model to static input scans of subjects captured with and without loose clothing. In this scenario, we compare the accuracy achieved by our method to that of the variant of the commonly used SCAPE model proposed by Jain et al.~\cite{Jain_etal_10_movie_reshape}. To simplify the presentation, we slightly abuse the notation and refer to this variant as SCAPE model in the following. We used the MPI database~\cite{HasStoSunRosSei09} to learn the SCAPE model using all models in standard posture for the shape model and using a single model in 35 postures for the posture model. For the shape model, $97\%$ of the variability present in the training data are retained, as we observe empirically that over-fitting does not occur when learning from this database of models in standard posture. For the SCAPE fitting, a constrained optimization is used to find the shape and posture parameters located within three standard deviations of the model mean. The SCAPE fitting iteratively fits to nearest neighbors.

To train our model, we use the scans of all subjects in all available postures of the MPI database. That is, for the same training database, our method is able to leverage more scans for training. For all experiments shown in this section, the 14 landmarks $l_j$ are picked manually and provided as input to both fitting algorithms. 

\begin{figure}[tb]
\centering
\includegraphics[width=12.0cm]{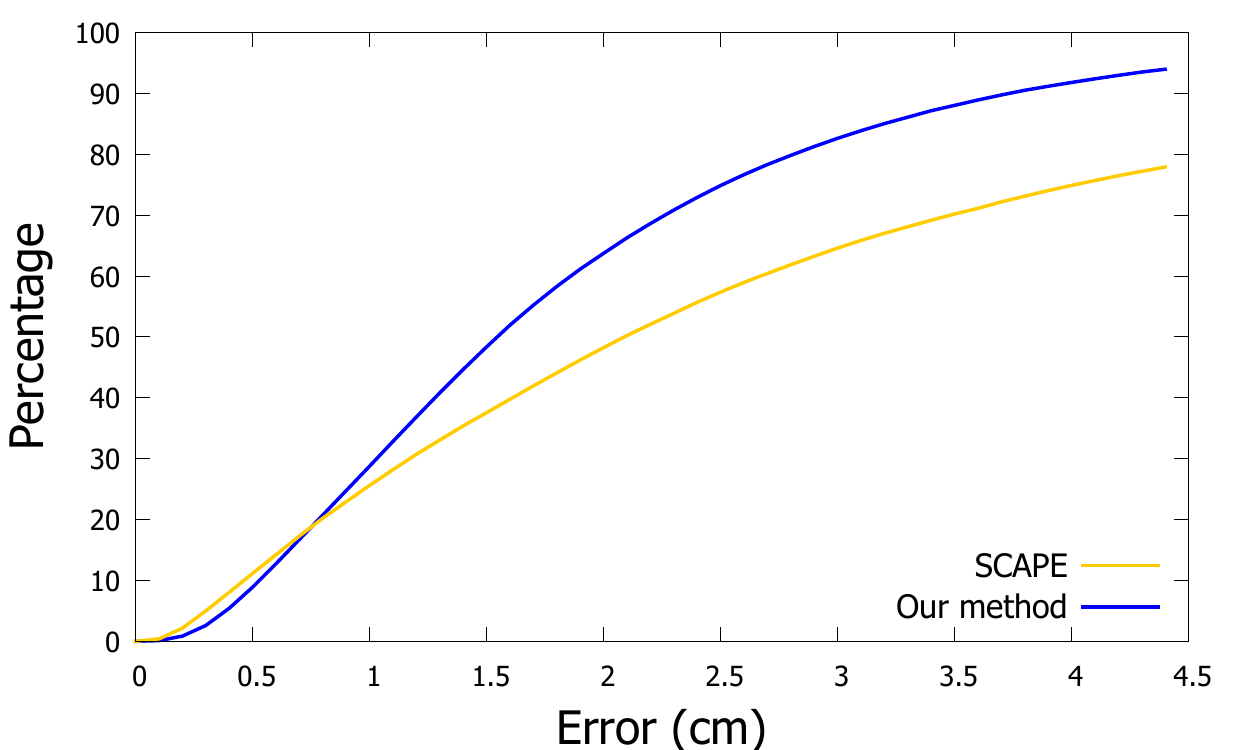}
\caption{Results of fitting SCAPE and our model to a subset of the MPI database.}
\label{fig:results_MPI_DB}
\end{figure}

\paragraph{Subjects in minimal tight clothing} We first show an experiment, where we aim to fit the statistical model to input scans representing subjects in minimal tight clothing. To evaluate the fitting accuracy in this case, we divided the subjects in the MPI database into two halves. We used one half to train both the SCAPE model and our model (the 260 scans of subjects $1-53$), and the other half was used for testing (the 260 scans of subjects $54-115$). For the SCAPE model, again all available shapes in standard posture were used for the shape model and a single subject in 35 postures was used for the posture model, while we used all available scans of half of the database to train our model. The two learned statistical models were then fitted to the remaining models of the database. Figure~\ref{fig:results_MPI_DB} shows the cumulative plots of the distances of the vertices of the fitting results to their corresponding vertices in the registered MPI database. Our method outperforms SCAPE. Some of the high errors for both methods stem from noise in the database. For SCAPE, many of the high errors are in the area of the torso, which is not always fitted well to the data as no landmarks are used to guide the model in this area, and as consequently, the posture model learned by SCAPE fails to fit accurately to the data. In contrast, our skeleton-based posture fitting usually fits the model well to the data in spite of the lack of landmarks in the torso area.

\begin{figure}[tb]
\centering
\begin{tabular}{c c c c c c}
\includegraphics[height=3.0cm]{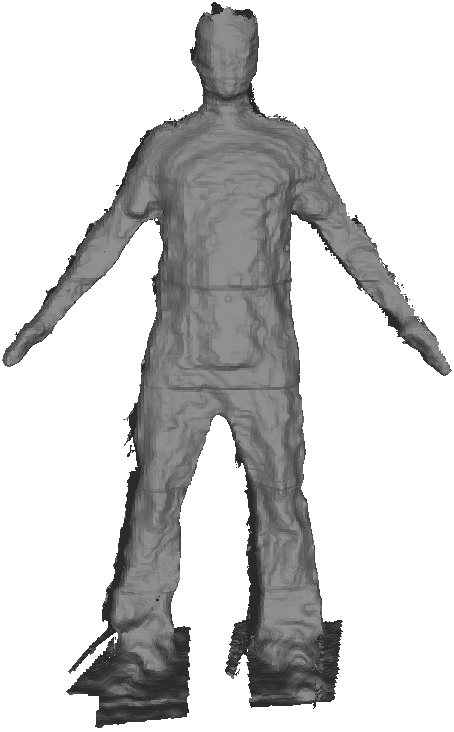} &
\includegraphics[height=3.0cm]{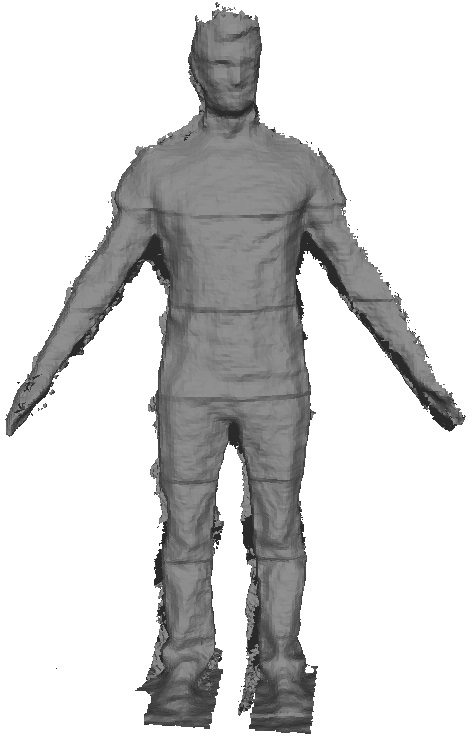} &
\includegraphics[height=3.0cm]{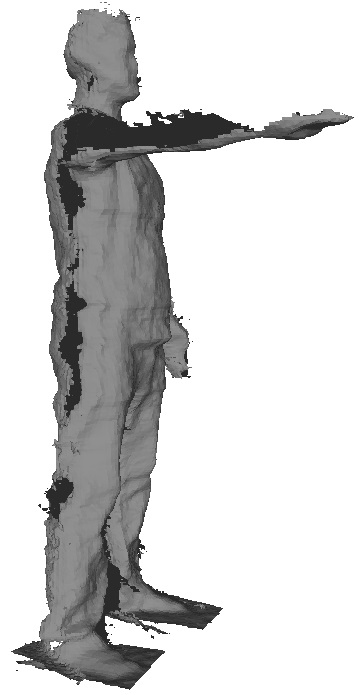} &
\includegraphics[height=3.0cm]{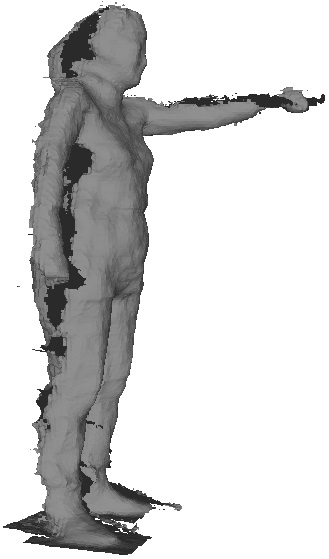} &
\includegraphics[height=3.0cm]{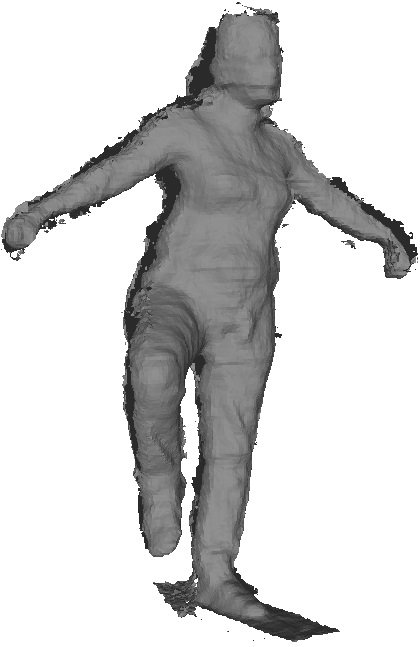} &
\includegraphics[height=3.0cm]{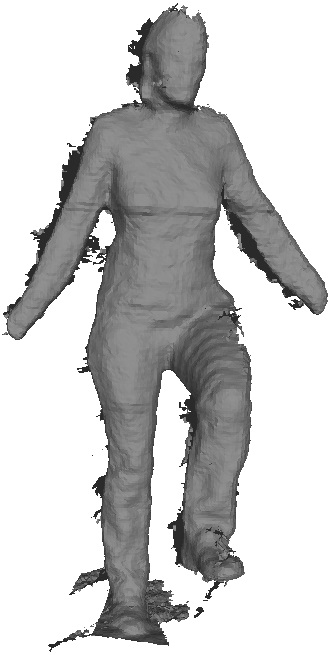} \\
\includegraphics[height=3.0cm]{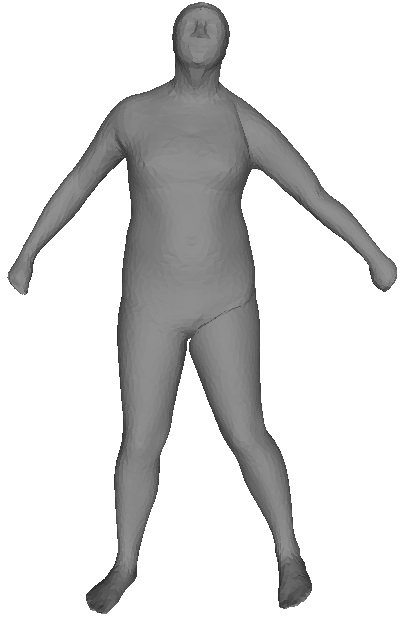} &
\includegraphics[height=3.0cm]{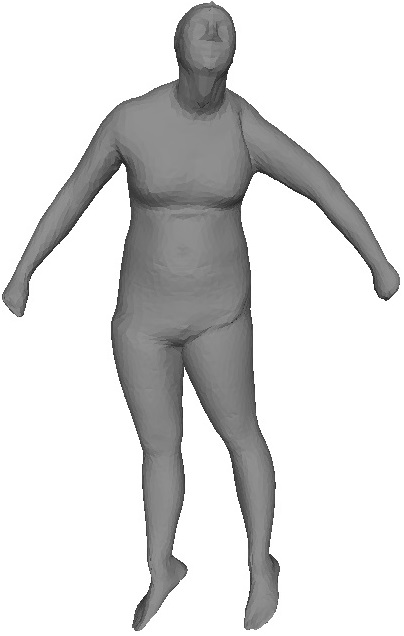} &
\includegraphics[height=3.0cm]{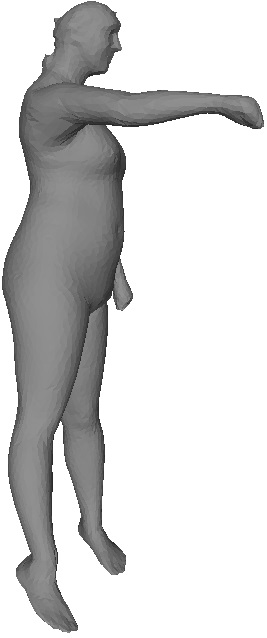} &
\includegraphics[height=3.0cm]{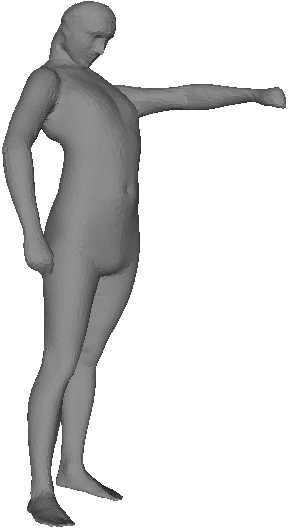} &
\includegraphics[height=3.0cm]{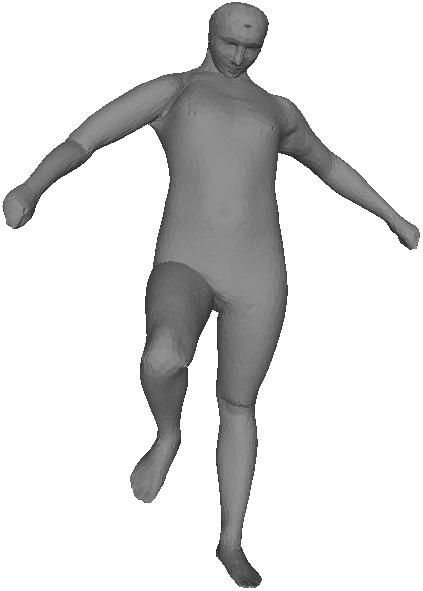} &
\includegraphics[height=3.0cm]{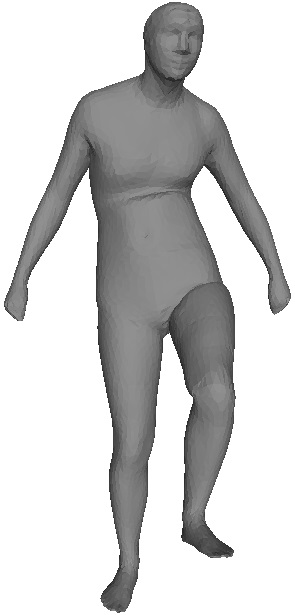} \\
\includegraphics[height=3.0cm]{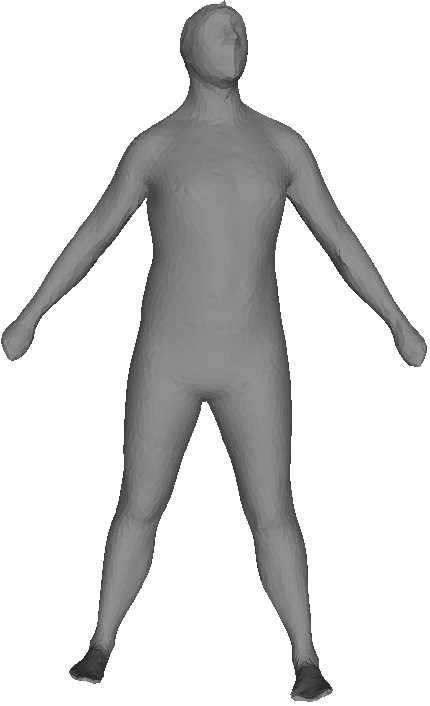} &
\includegraphics[height=3.0cm]{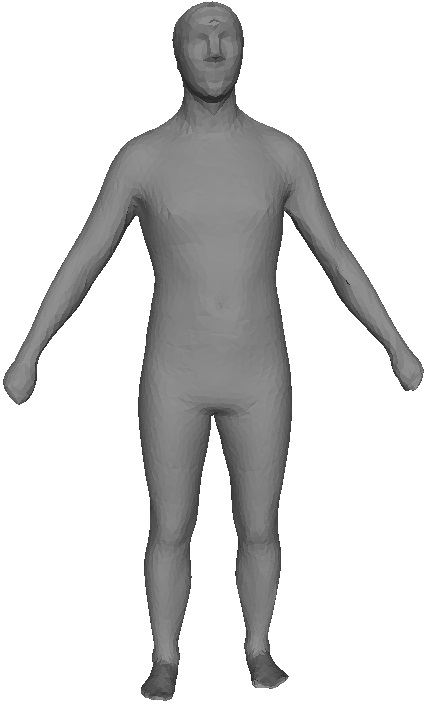} &
\includegraphics[height=3.0cm]{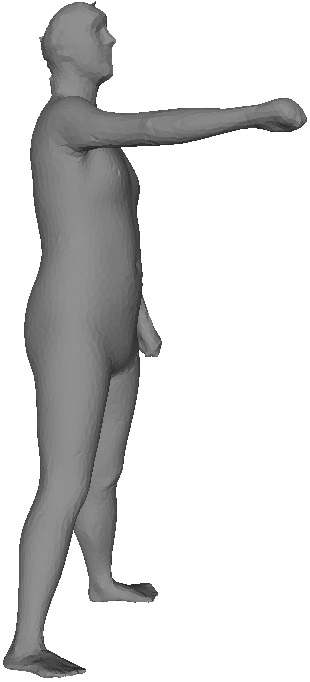} &
\includegraphics[height=3.0cm]{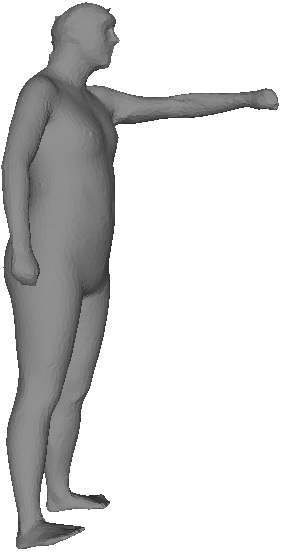} &
\includegraphics[height=3.0cm]{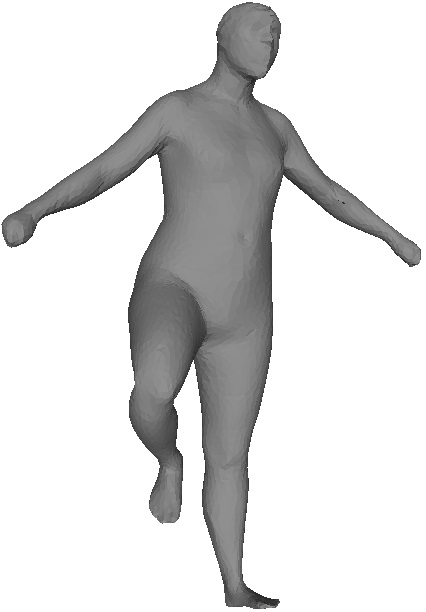} &
\includegraphics[height=3.0cm]{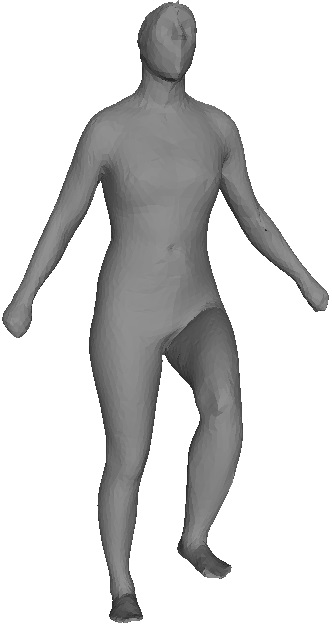} \\
\end{tabular}
\caption{Top: data set of static scans of people dressed in regular clothing. Middle: results of fitting SCAPE model to a single scan. Bottom: results of fitting our model to a single scan.}
\label{fig:kinect_static}
\end{figure}

\paragraph{Subjects in casual clothing} To evaluate our algorithm on a database of more challenging static scans, we collected a data set consisting of a total of 18 body scans of 4 subjects dressed in regular casual office clothing in up to 5 postures each using Kinect Fusion~\cite{Izadi2011}. We simultaneously captured the front and back views for every subject in each posture separately using Kinect Fusion and manually merged the two resulting views. Some of the scans (covering all 4 subjects and 5 postures) are shown in the first row of Figure~\ref{fig:kinect_static}. The postures were chosen to resemble the postures used by Balan and Black~\cite{BalanB08}. We could not use their data directly, as their method takes a small set of input images (not covering the full view of the body), while we require a scan that covers the full body. Note that the scans are corrupted by noise and missing data. For each of the four subjects, we further recorded the height, waist circumference, and chest circumference. We use these measures to evaluate the accuracy of the fitting results by computing the corresponding measurements on the resulting fitted models. 

For both SCAPE and our method, we perform two ways of data fitting. First, we fit the models to each input scan (in a single posture) individually, and second, we fit the models to all postures available for a given subject jointly by solving for a single body shape estimate and multiple posture estimates. 

\begin{figure}[tb]
\centering
\includegraphics[width=10.5cm]{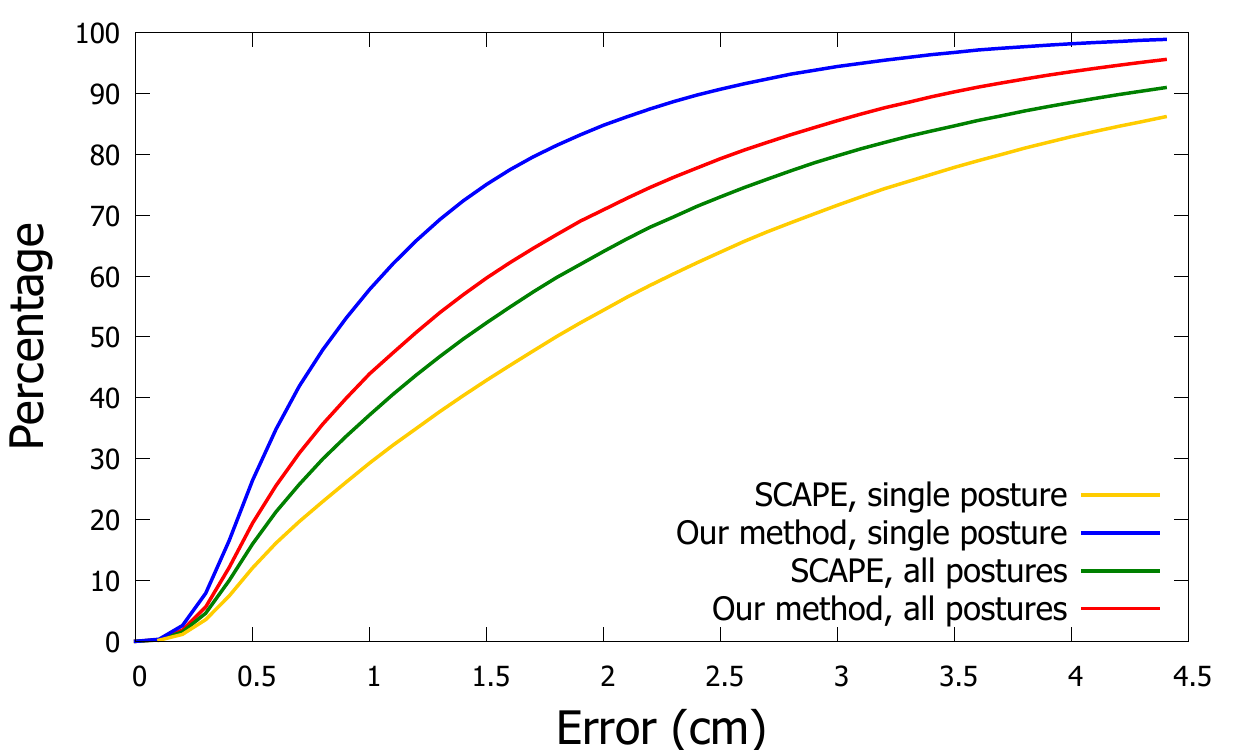}
\caption{Cumulative distances when fitting SCAPE and our model to a data set of 18 scans acquired using Kinect Fusion.}
\label{fig:results_Kinect_fit}
\end{figure}

\begin{figure}[tb]
\centering
\includegraphics[width=10.0cm]{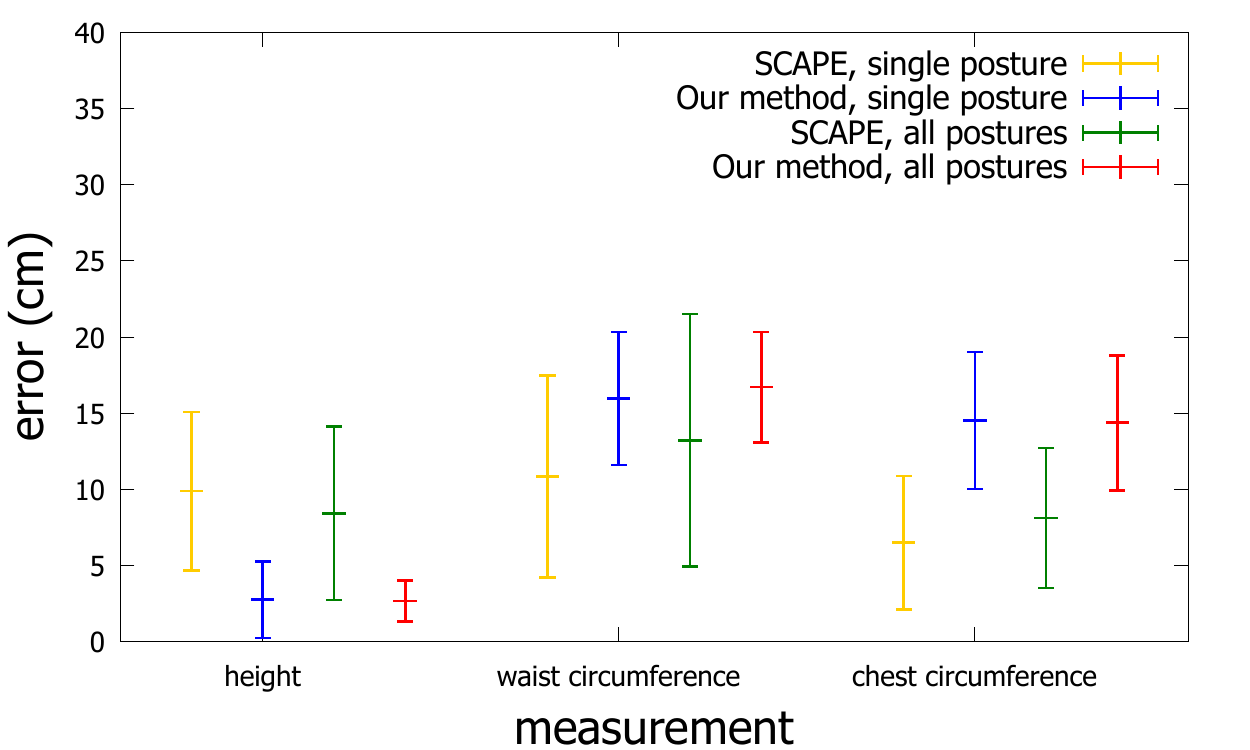}
\caption{Measurement errors of estimate by SCAPE and our model of a data set of 18 scans acquired using Kinect Fusion. Plot shows means and standard deviations of the errors.}
\label{fig:results_Kinect_DB}
\end{figure}

To evaluate the results, we first measure the fitting accuracy by computing the distance between each vertex of the result and its closest point on the input data. Figure~\ref{fig:results_Kinect_fit} summarizes the fitting accuracy. Note that for both options, our method leads to models that are closer to the input data than SCAPE. For our method, the distance to the input data increases when multiple postures are fitted simultaneously. This is to be expected as multiple observations of a dressed subject give more cues about the body shape, which leads to a better body shape estimate that may deviate more from the data, which includes details of clothing. To see that our body shape estimate improves when multiple postures are used, refer to Figure~\ref{fig:results_Kinect_DB} (discussed in detail below), where the improvements can be seen from the reduced standard deviations, which is especially visible for the height measurement. For the SCAPE model, the opposite behaviour can be observed. The reason is that the SCAPE model is not fitted well to a single input scan, as can be seen in Figure~\ref{fig:kinect_static}, which shows some fitting results. Note that the results using our method represent realistic body shapes and postures that are close to the input scans, while this is not always the case for the results using SCAPE. For instance, the following body parts are estimated inaccurately in the results found by SCAPE: the posture of the legs shown in the second column, the posture of the feet shown in the third column, the posture of the upper back shown in the fourth column, and the posture of the head shown in the fifth column.

Second, we measure the height and circumferences on the fitting results. The circumference measurements are computed by intersecting the torso of the model with a plane parallel to the floor plane and by computing the length of the convex hull of this intersection. The results for the different methods are summarized in Figure~\ref{fig:results_Kinect_DB}. While our method predicts the height of the models quite accurately (even though some of the subjects wore shoes during acquisition), the waist and chest circumferences are overestimated because the clothing tricks the method into predicting body shapes with larger circumferences. This is especially true for the waist circumference, where the body shape of the acquired subjects is hidden by large clothing folds, as can be seen in Figure~\ref{fig:kinect_static}. SCAPE leads to a significantly worse estimate of the height, but to better estimates of the circumferences. Note however that while the two estimated circumferences have low error for SCAPE, the estimated body shape is often inaccurate, as can be seen in the chest area of the model shown in the first column of Figure~\ref{fig:kinect_static}. Here, the overall body shape estimate of our method is closer to the input data than the one by SCAPE. Furthermore, the estimates of the circumferences found using SCAPE get worse when the model is fitted to multiple scans simultaneously. The reason is that using multiple scans leads to fitting results that are closer to the data (as can be seen in Figure~\ref{fig:results_Kinect_fit}), which leads to overestimated circumferences due to the clothing. 

\begin{figure}[tb]
\centering
\includegraphics[width=10.0cm]{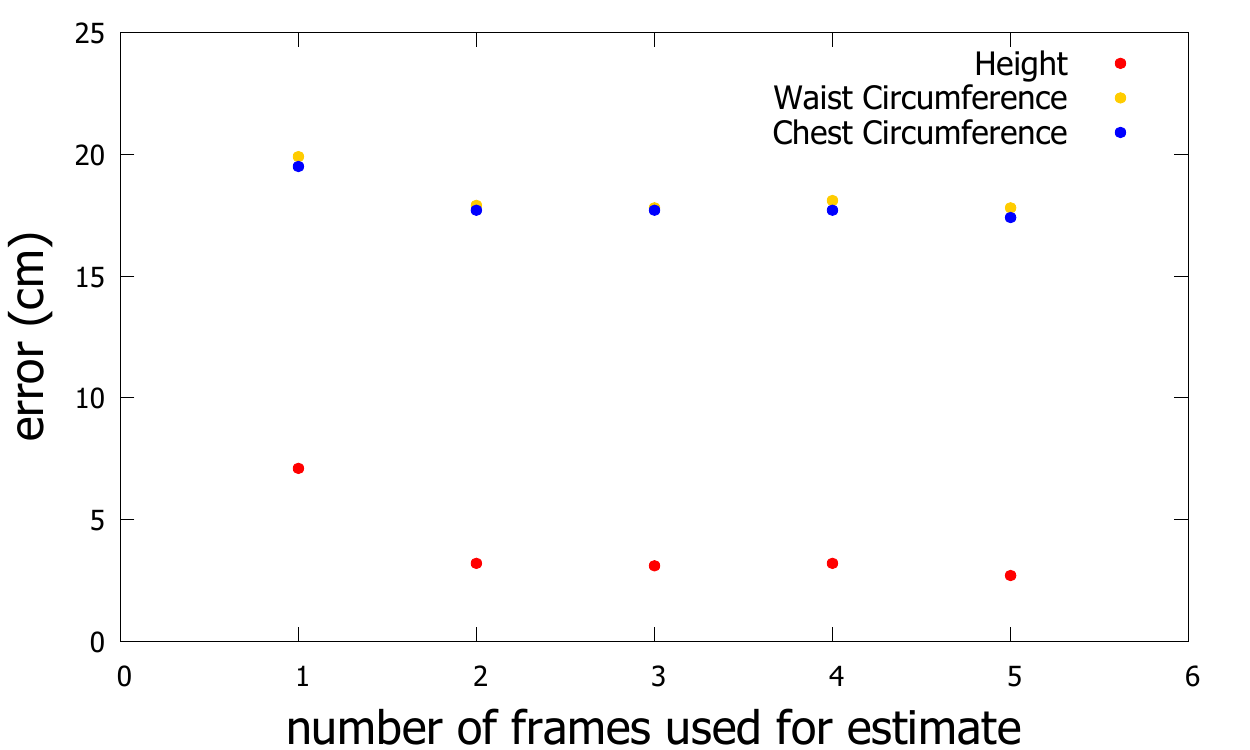}
\caption{Errors of estimated height, waist circumference, and chest circumference measurements of the scan shown in the leftmost column of Figure~\ref{fig:kinect_static} with increasing number of scans used to estimate the body shape.}
\label{fig:results_Kinect_s1p1}
\end{figure}

When fitting multiple postures simultaneously using our method, the standard deviations of all measurements decrease, which indicates that the errors get spread more evenly, which is to be expected as the shapes are averaged in the shape space $\mathcal{S}$. We observed that for some scans, the measurement errors decrease significantly, while for other scans, there is a slight increase in some of the measurement errors. Having additional observations mainly improves the accuracy of the shape estimate for frames that had a high error with single frame fitting compared to the other available frames. One instance where the errors decrease significantly is the scan shown in the leftmost column of Figure~\ref{fig:kinect_static}. Here, the error on the height, waist, and chest measurements decrease by 4.4cm, 2.1cm, and 2.1cm, respectively, by using all postures instead of a single one. Figure~\ref{fig:results_Kinect_s1p1} shows how the measurement error decreases when increasing the number of scans used to estimate the body shape from one to five. Note that all errors decrease significantly when using two scans instead of one to compute the shape estimate, while additional frames only lead to minor improvements.

\begin{figure}[tb]
\centering
\footnotesize
\begin{tabular}{|c|c|c|c|c|c|}
\hline
Frame \# & & 1 & 6 & 11 & 16 \\
\hline
\hline
\multirow{2}{*}[1.2cm]{No Noise} & \multirow{1}{*}[1.2cm]{Input} & \includegraphics[scale=0.1]{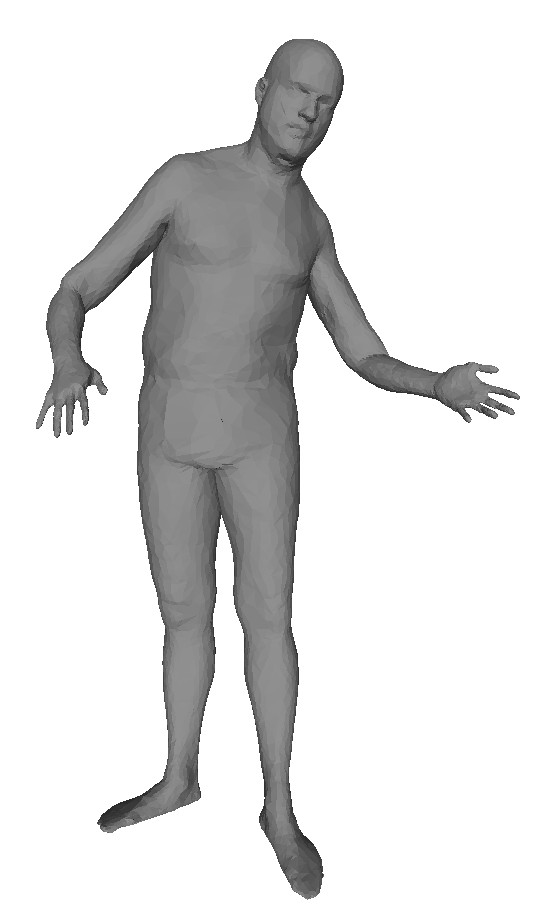} & \includegraphics[scale=0.1]{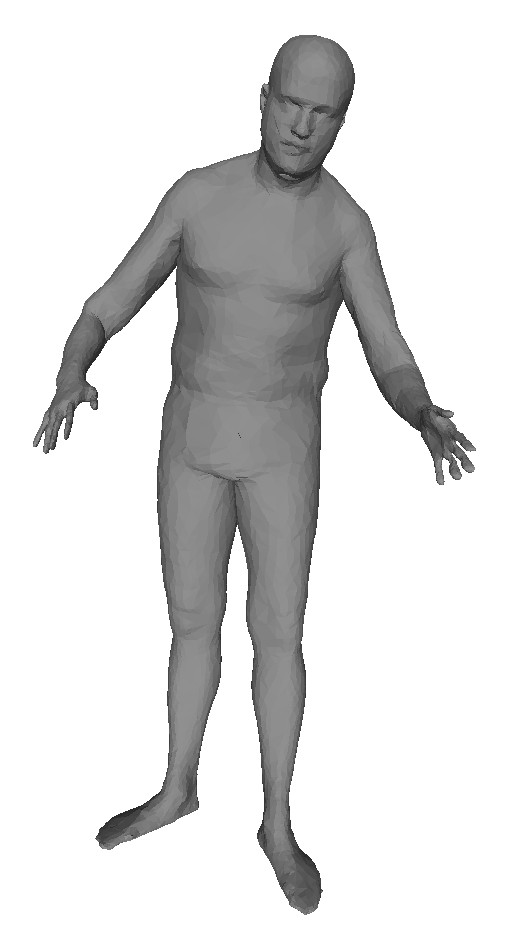} & \includegraphics[scale=0.1]{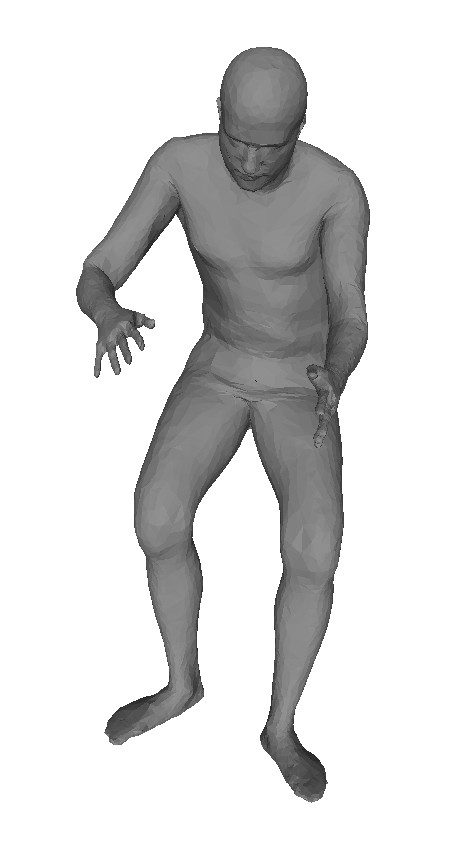} & \includegraphics[scale=0.1]{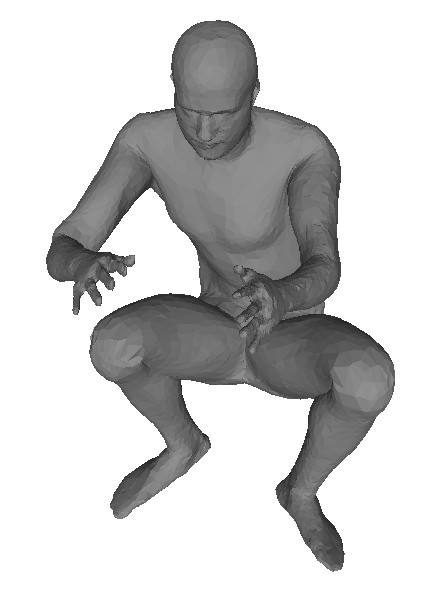} \\
\cline{2-6}
 & \multirow{1}{*}[1.2cm]{Result} & \includegraphics[scale=0.1]{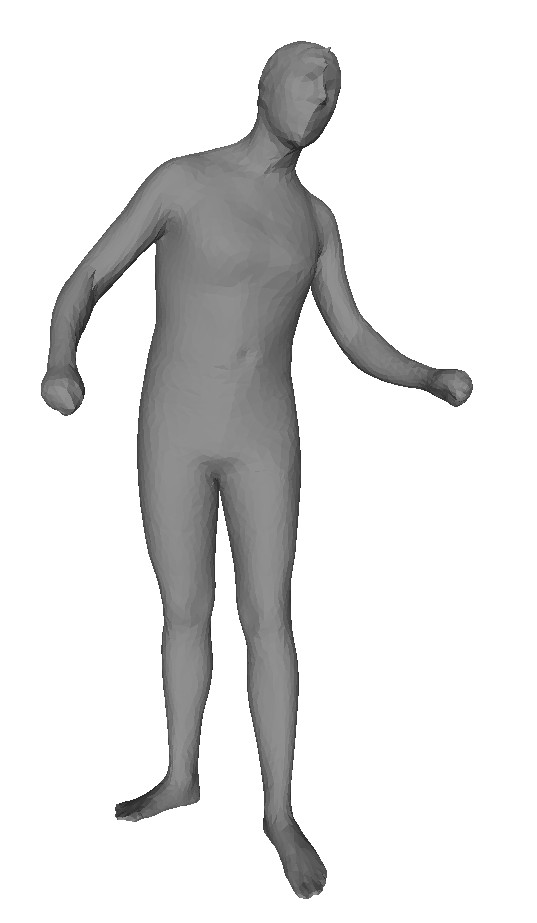} & \includegraphics[scale=0.1]{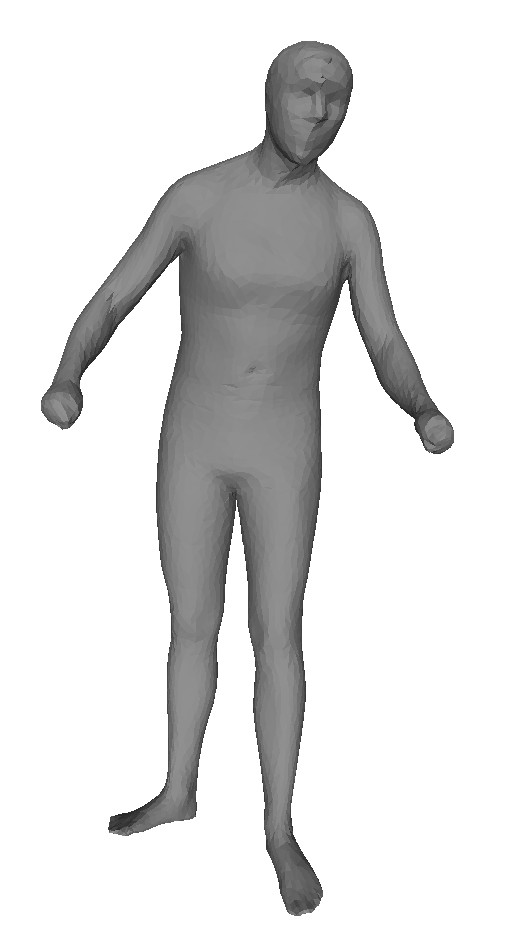} & \includegraphics[scale=0.1]{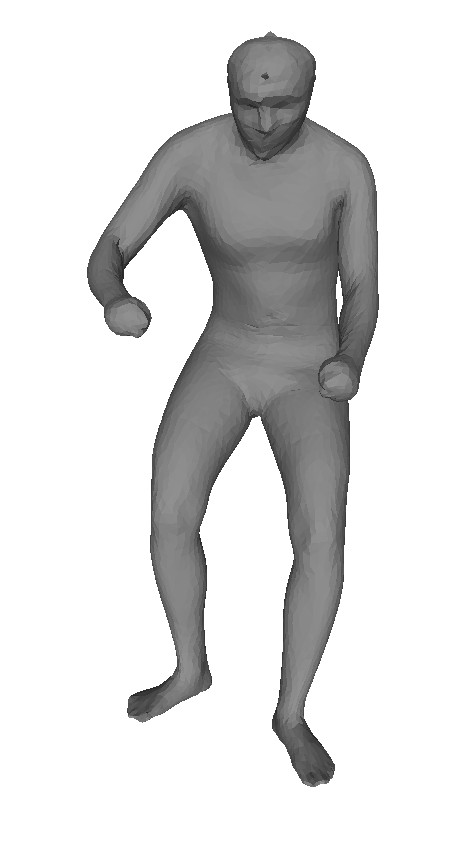} & \includegraphics[scale=0.1]{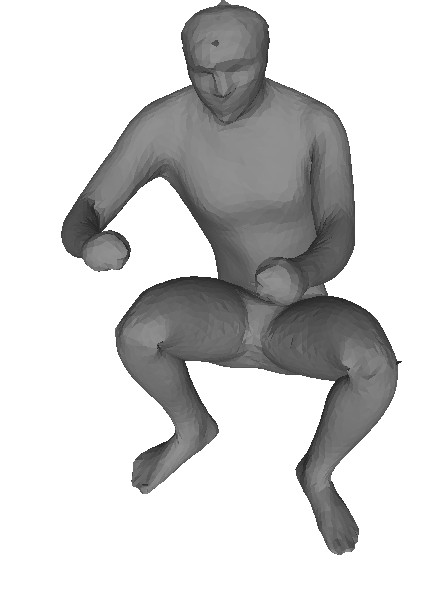} \\
\hline
\hline
\multirow{2}{*}[1.2cm]{Gaussian Noise} & \multirow{1}{*}[1.2cm]{Input} & \includegraphics[scale=0.1]{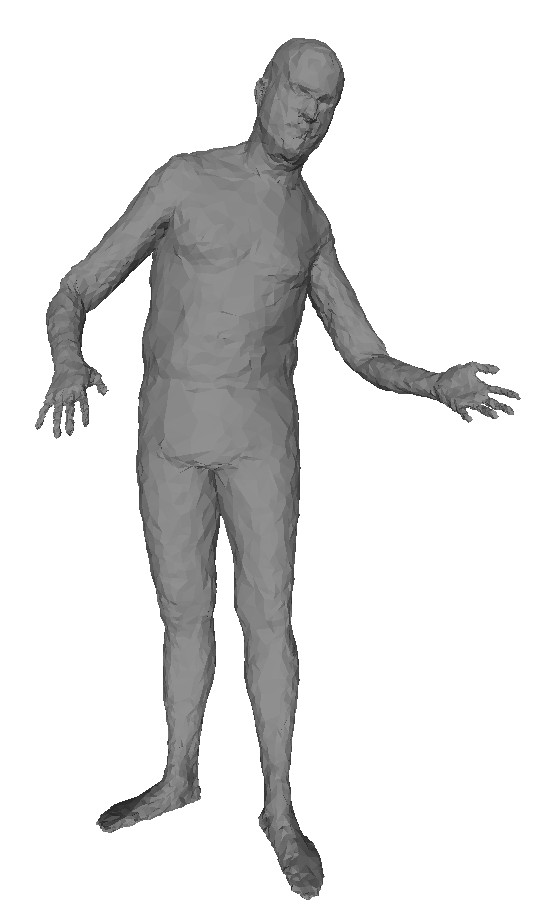} & \includegraphics[scale=0.1]{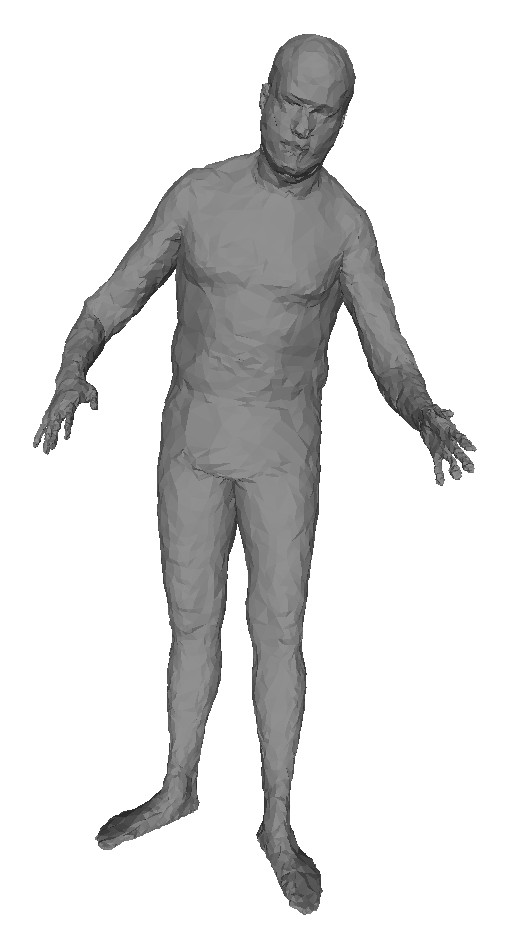} & \includegraphics[scale=0.1]{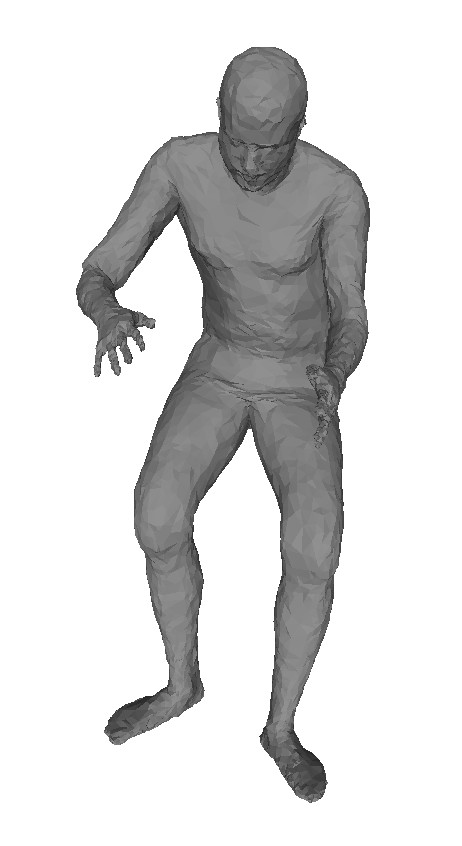} & \includegraphics[scale=0.1]{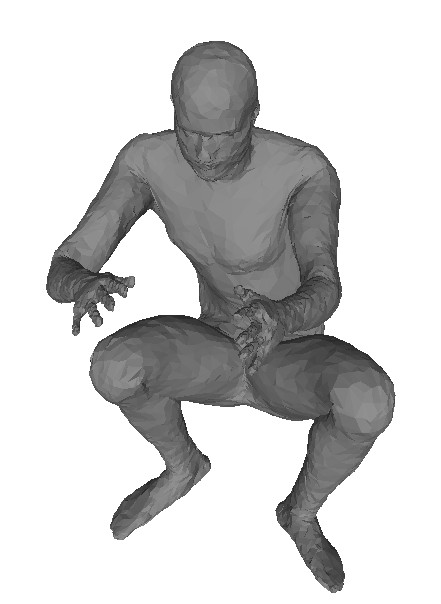} \\
\cline{2-6}
 & \multirow{1}{*}[1.2cm]{Result} & \includegraphics[scale=0.1]{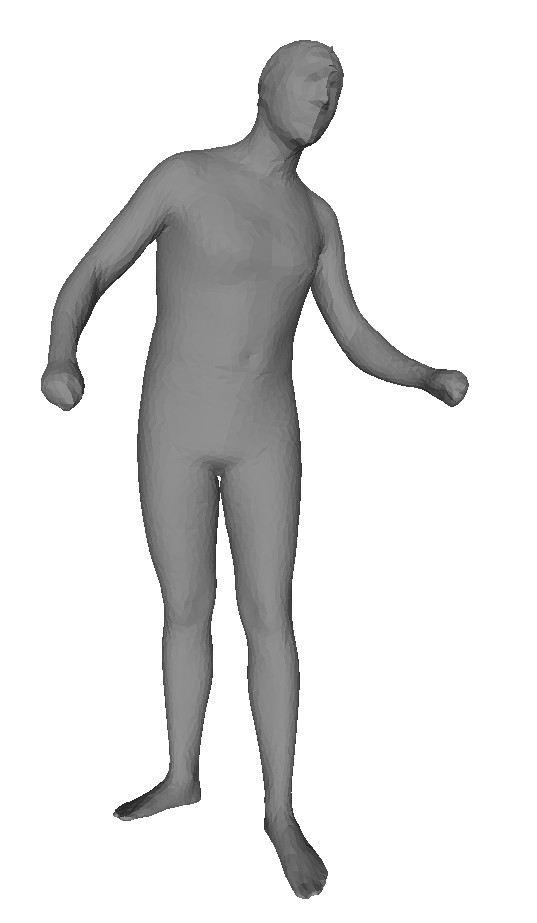} & \includegraphics[scale=0.1]{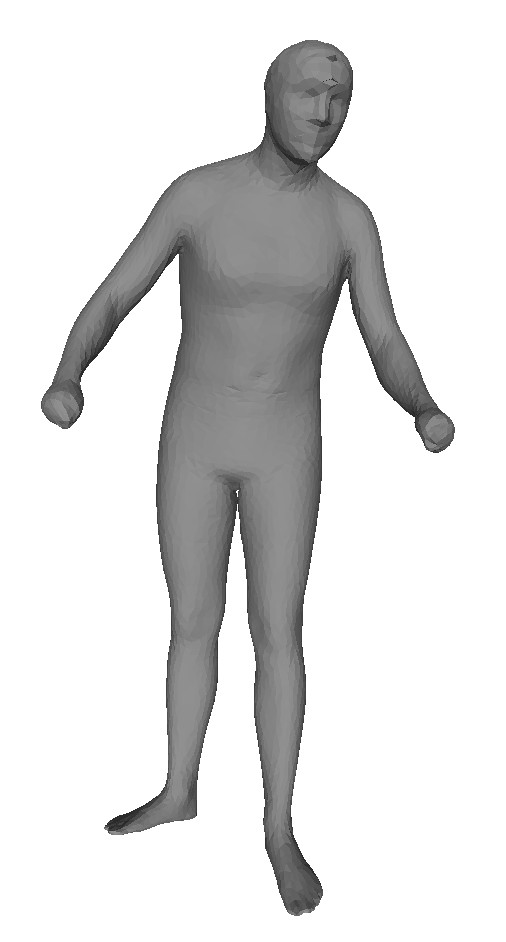} & \includegraphics[scale=0.1]{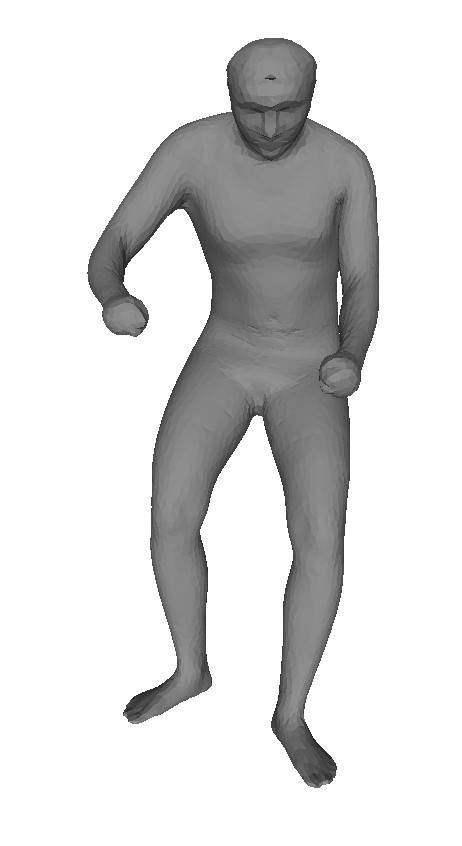} & \includegraphics[scale=0.1]{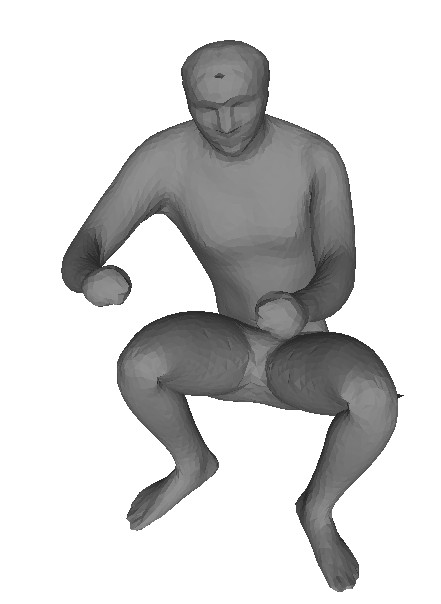} \\
\hline
\hline
\multirow{2}{*}[1.2cm]{Outliers} & \multirow{1}{*}[1.2cm]{Input} & \includegraphics[scale=0.1]{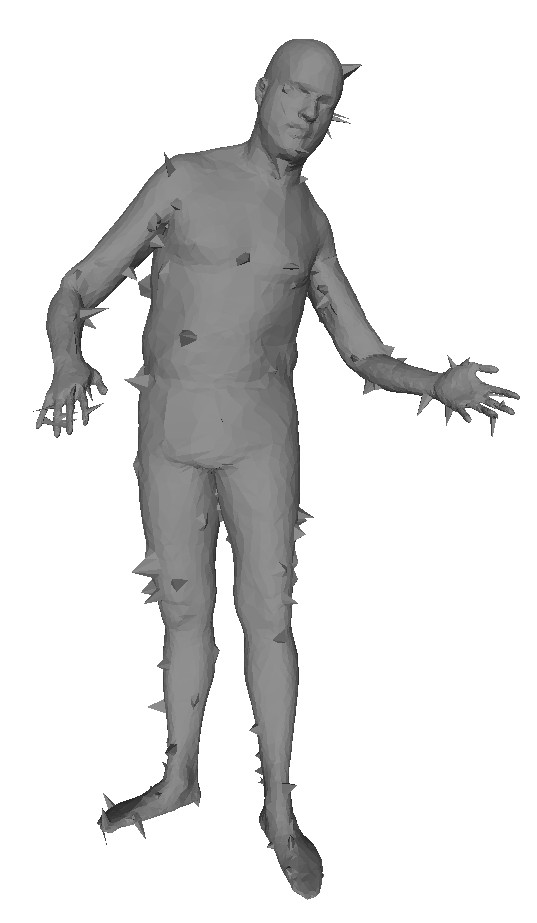} & \includegraphics[scale=0.1]{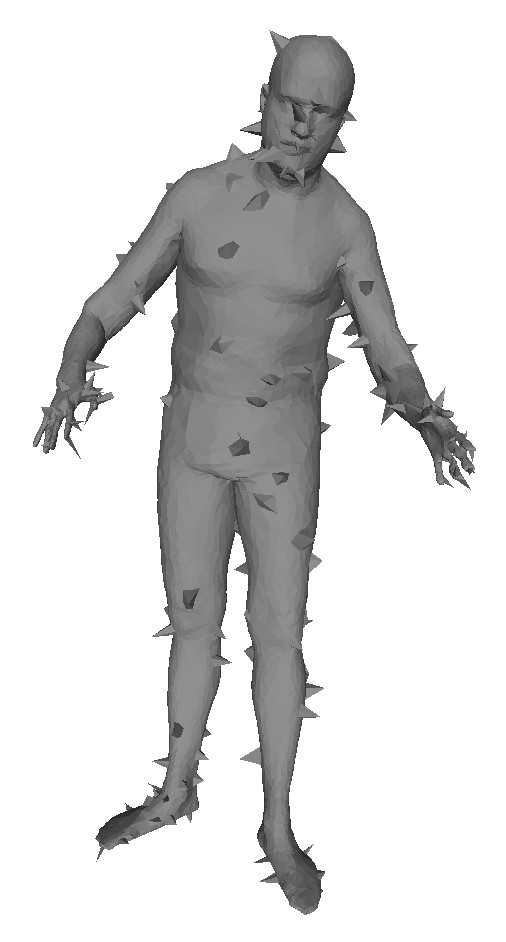} & \includegraphics[scale=0.1]{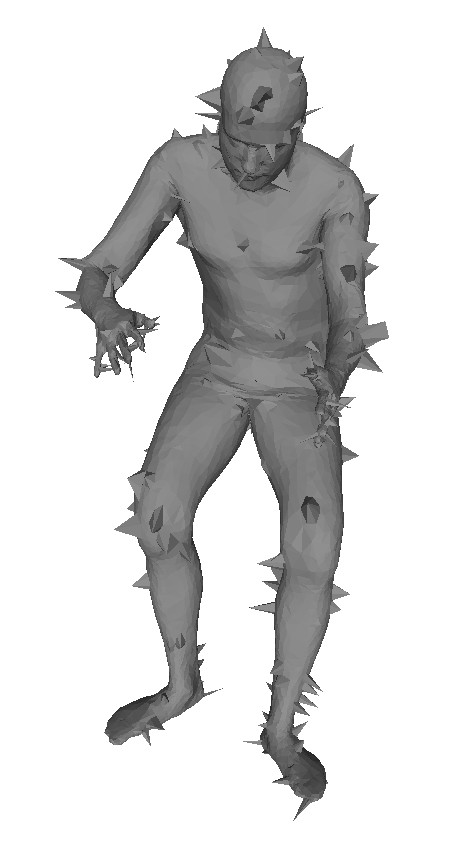} & \includegraphics[scale=0.1]{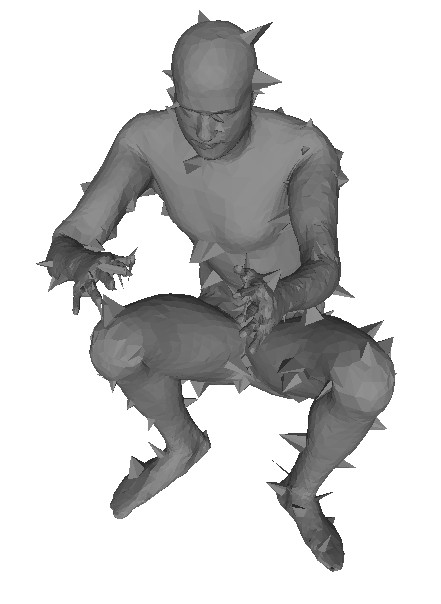} \\
\cline{2-6}
 & \multirow{1}{*}[1.2cm]{Result} & \includegraphics[scale=0.1]{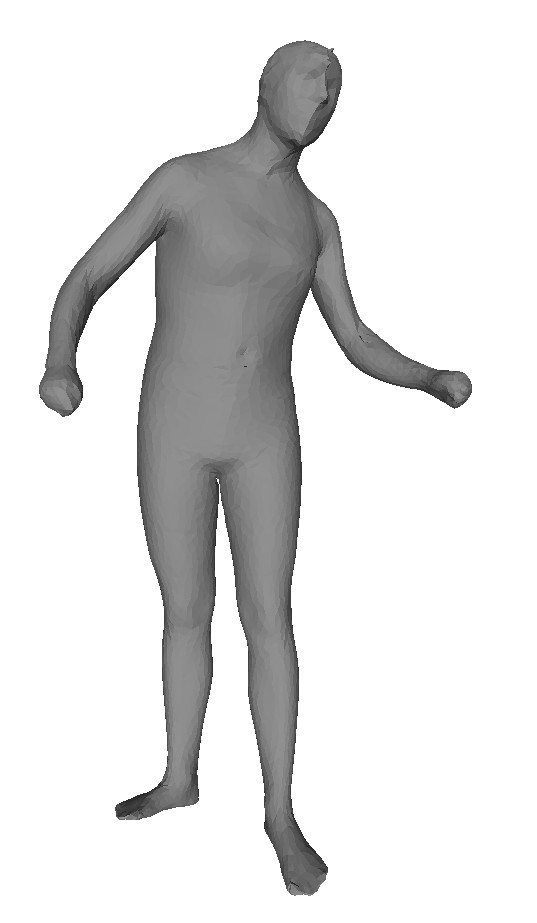} & \includegraphics[scale=0.1]{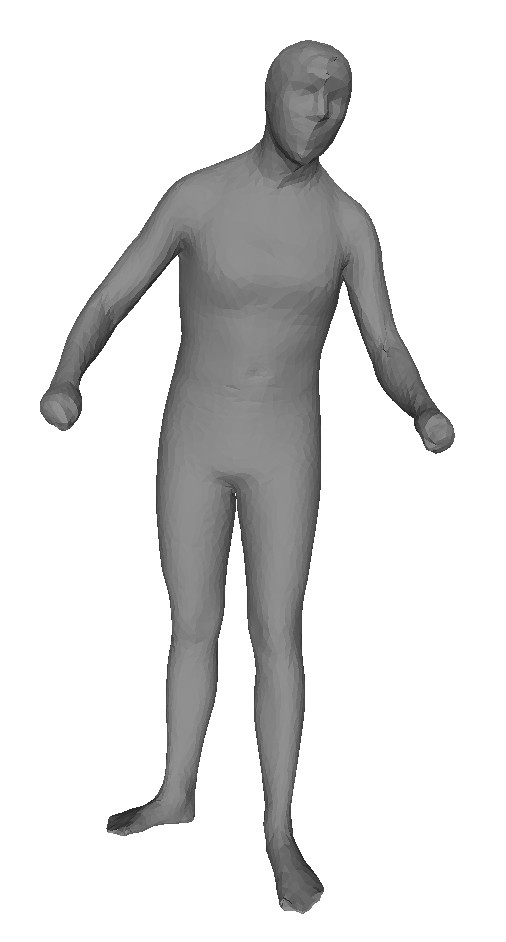} & \includegraphics[scale=0.1]{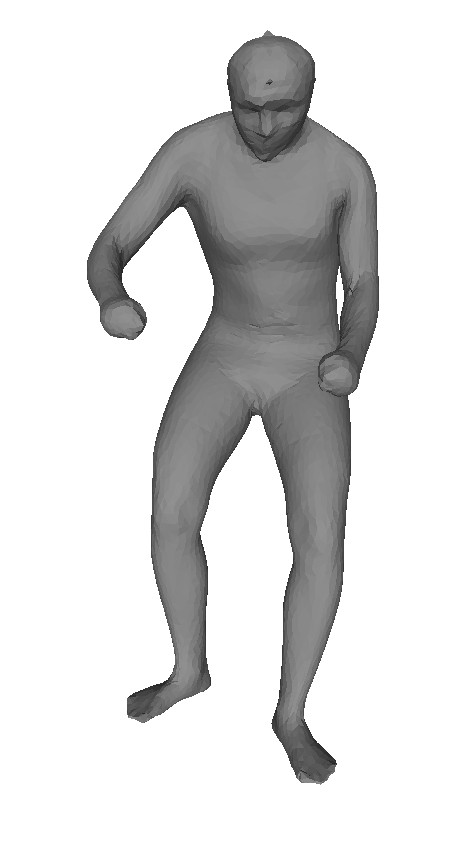} & \includegraphics[scale=0.1]{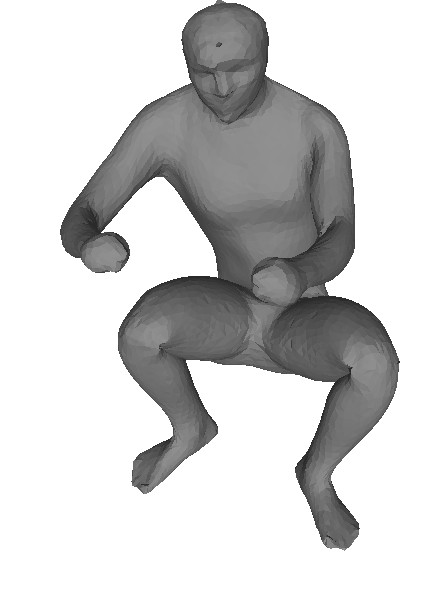} \\
\hline
\hline
\multirow{2}{*}[1.2cm]{Outliers} & \multirow{1}{*}[1.2cm]{Input} & \includegraphics[scale=0.1]{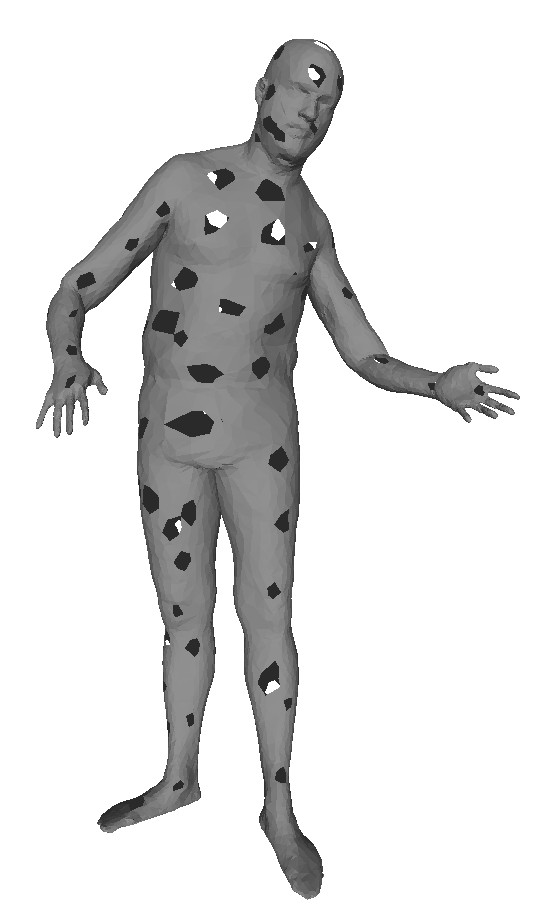} & \includegraphics[scale=0.1]{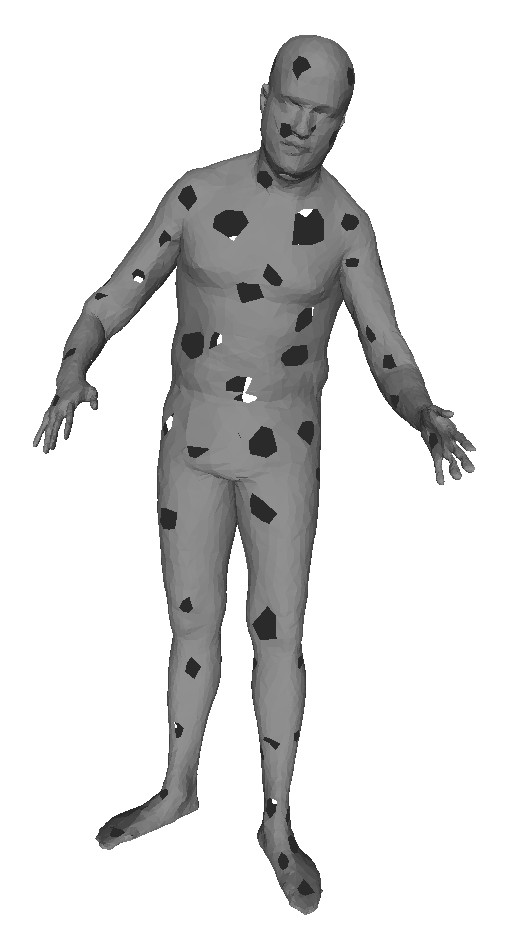} & \includegraphics[scale=0.1]{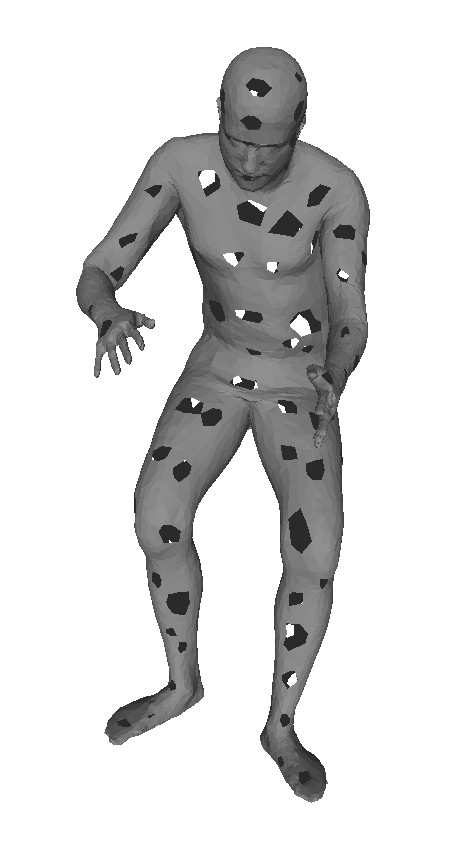} & \includegraphics[scale=0.1]{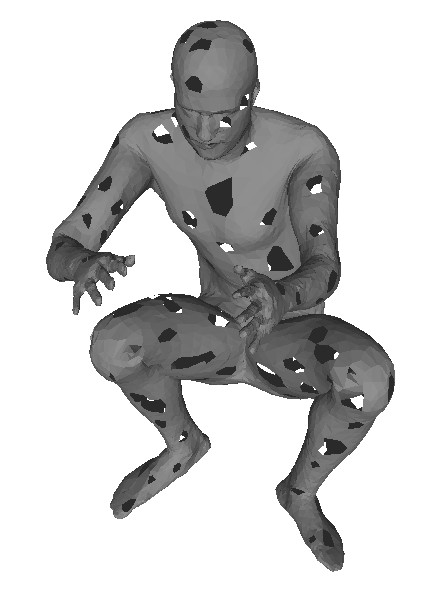} \\
\cline{2-6}
 & \multirow{1}{*}[1.2cm]{Result} & \includegraphics[scale=0.1]{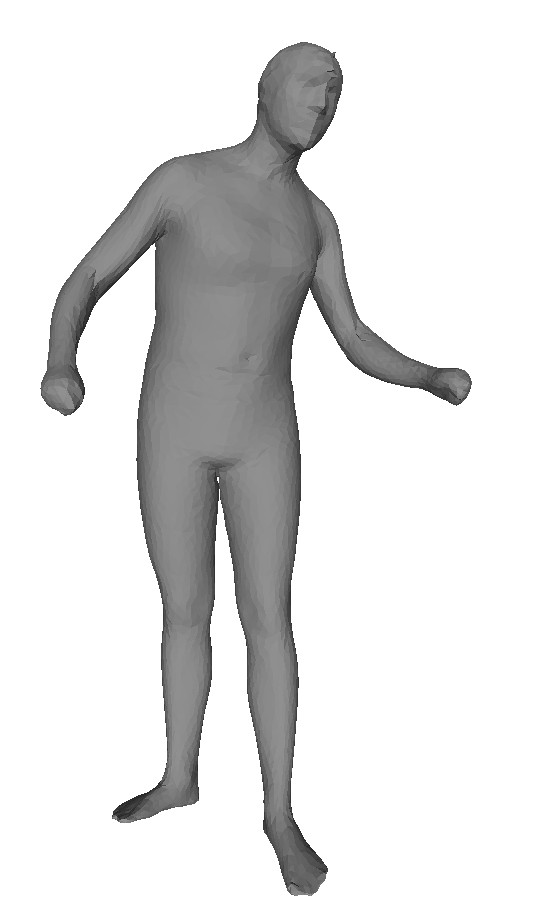} & \includegraphics[scale=0.1]{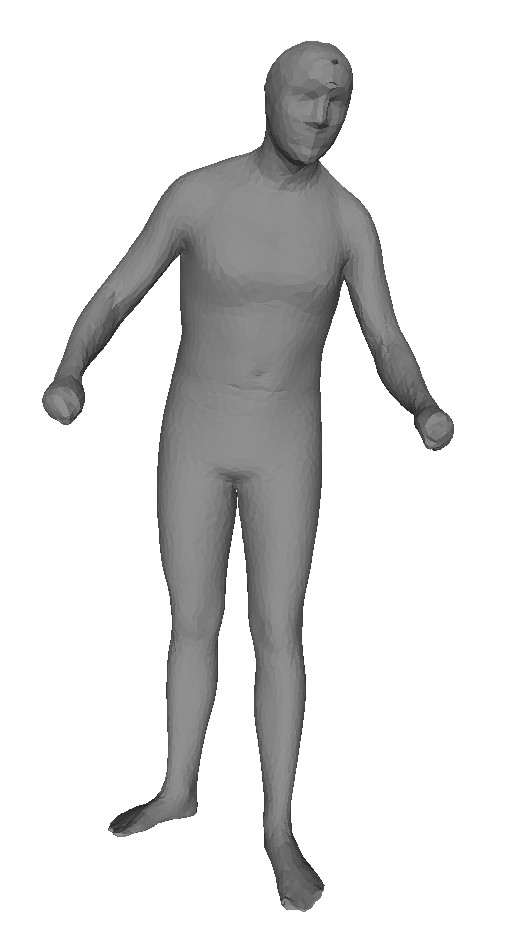} & \includegraphics[scale=0.1]{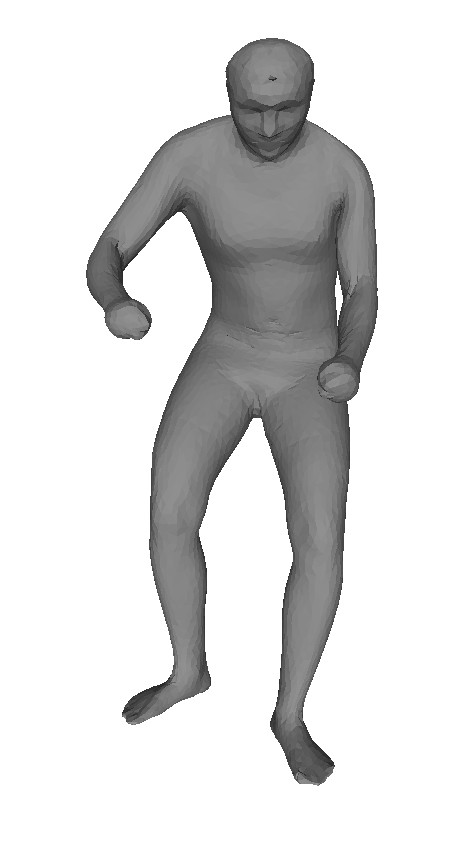} & \includegraphics[scale=0.1]{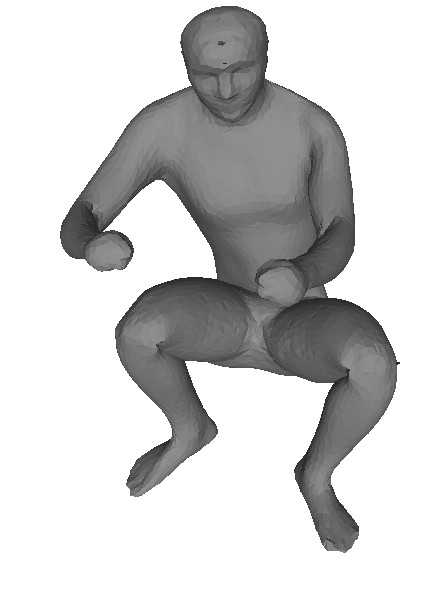} \\
\hline
\end{tabular}
\normalsize
\caption{Synthetic noise evaluation. Each row shows the input data and the results of our method.}
\label{fig:results_artificial}
\end{figure}
\clearpage

To conclude, we showed that for the fitting results to the 18 dressed subjects, our method leads to results that represent the overall body shape and posture correctly, while this is not always the case for SCAPE. Furthermore, the results found by our method are closer to the input data than the results found by SCAPE. While two circumference measurements are estimated more accurately using SCAPE than using our method, the overall body shapes predicted using SCAPE are often visually far from the true body shape. Hence, overall, the fitting accuracy of our method is higher than that of SCAPE. 

\subsection{Tracking Motion Sequences}

Next, we evaluate our method for tracking motion sequences showing humans with and without loose clothing. 

\begin{figure}[tb]
\centering
\includegraphics[width=12.0cm]{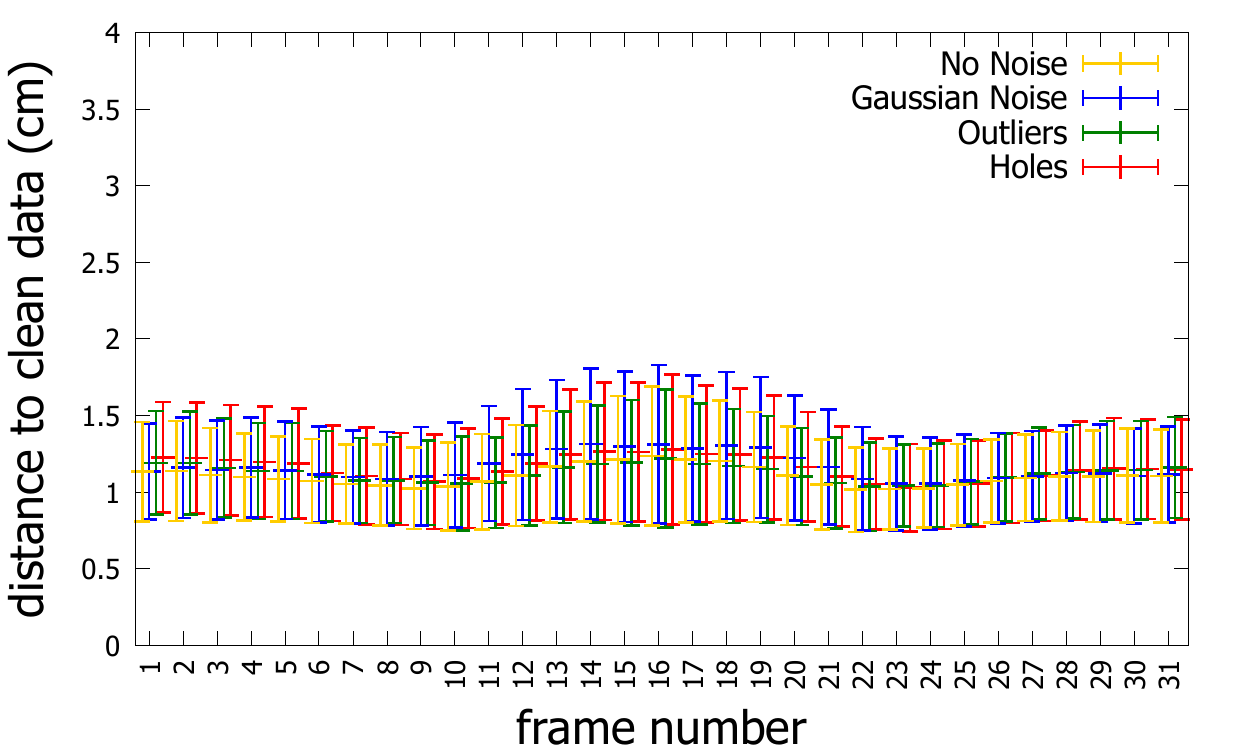}
\caption{Tracking of synthetic sequence corrupted by different types of noise. Plot shows means and standard deviations of the errors.}
\label{fig:evaluate_artificial}
\end{figure}

\paragraph{Synthetic motion sequences} We start by fitting our model to a synthetic motion sequence of a minimally dressed subject obtained by animating a processed scan of the CAESAR database~\cite{robinette_daanen_paquet_99_caesar} using Pinocchio~\cite{baran_popovic_07_animation}. This test allows to evaluate our method in the presence of controlled input noise. The following three types of noise are considered: (1) Gaussian noise with variance of $5\%$ of the bounding ball radius of the model applied to the input vertices, (2) outliers modeled by perturbing a vertex with probability $1/50$ along its normal direction by a magnitude that is uniformly distributed in the range $\left[0,4r\right]$, where $r$ is the average edge length of the model, and (3) holes that were added to the input models. For each sequence, we use our algorithm to track the data, and we evaluate the quality of the result by measuring the difference between the vertices on the result and their nearest neighbor in the original (uncorrupted) sequence. The model starts from a standing position, goes into a squatting position, and back to the starting standing position. Figure~\ref{fig:results_artificial} shows the input models and the results of the first half of the sequence, and Figure~\ref{fig:evaluate_artificial} shows the means and standard deviations of the distance of our result to the uncorrupted input model for each frame. The following two observations can be made. First, the tracking is stable, which means that there is no significant drift in the later frames. This can be seen as the motion is symmetric w.r.t. the squatting position (frame 16), and as frames corresponding to the same posture in the first and the second half of the motion sequence (i.e. frames $i$ and $32-i$ for $i=1,\ldots,15$) have similar error. This is due to the landmark prediction step that gives a good initialization to the posture fitting. Second, the synthetic noise does not have a significant influence on the results, which shows that our method is robust to different types of noise.

\begin{figure}[tb]
\centering
\centering
\footnotesize
\begin{tabular}{c c c c c c c}
\includegraphics[scale=0.12]{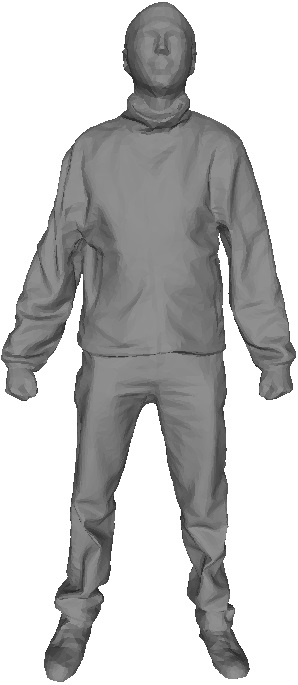} &
\includegraphics[scale=0.12]{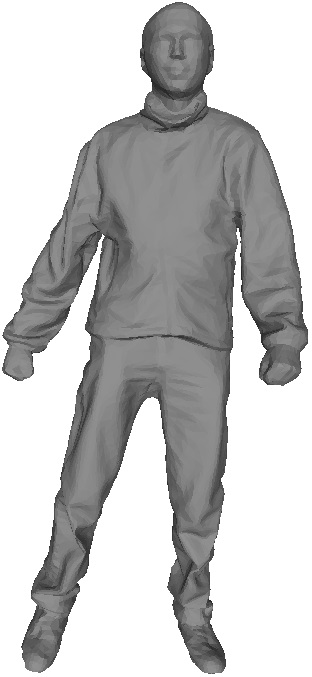} &
\includegraphics[scale=0.12]{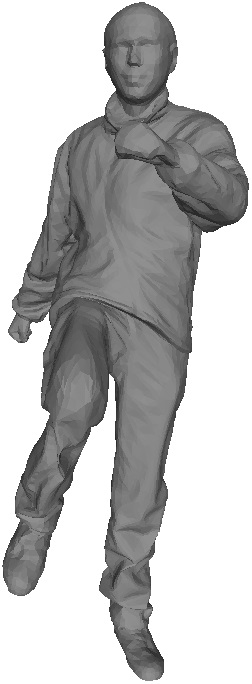} &
\includegraphics[scale=0.12]{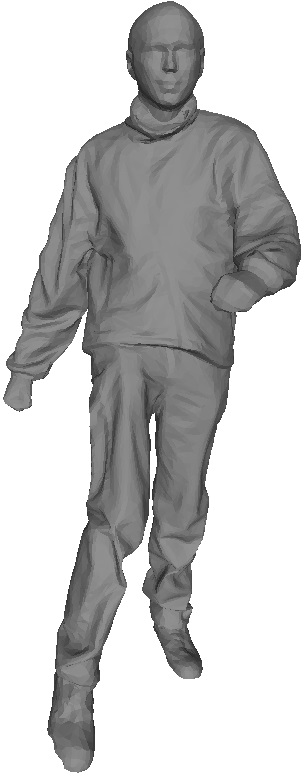} &
\includegraphics[scale=0.12]{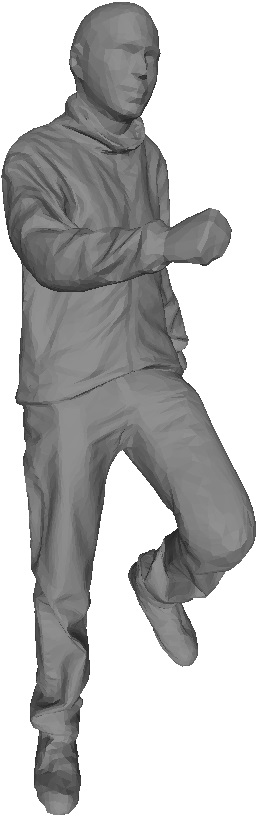} &
\includegraphics[scale=0.12]{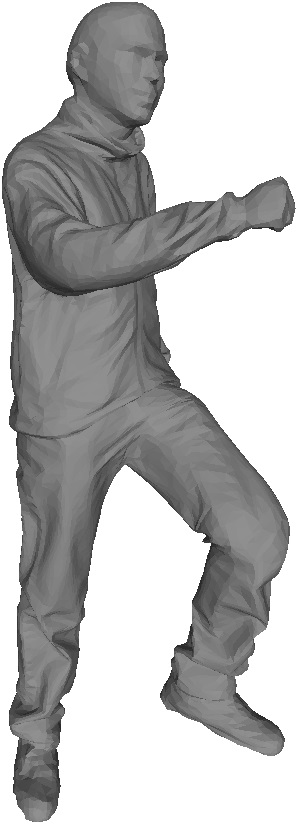} &
\includegraphics[scale=0.12]{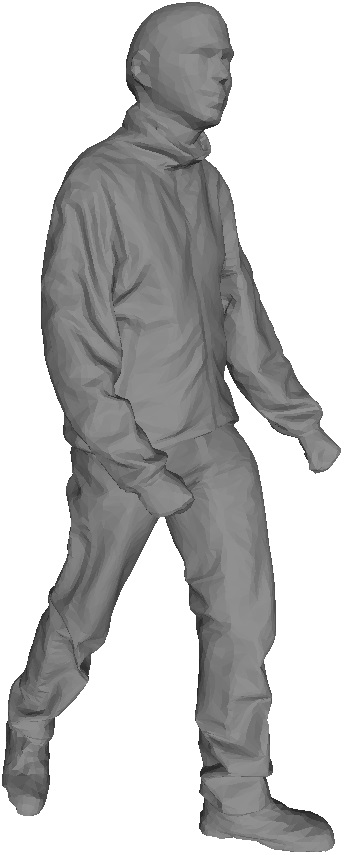} \\
\includegraphics[scale=0.12]{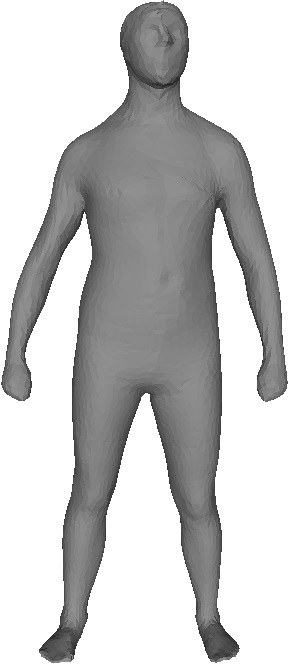} &
\includegraphics[scale=0.12]{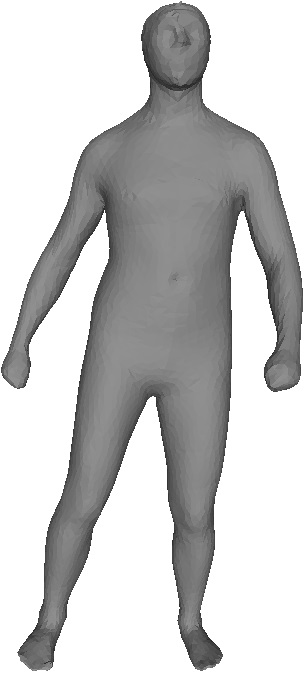} &
\includegraphics[scale=0.12]{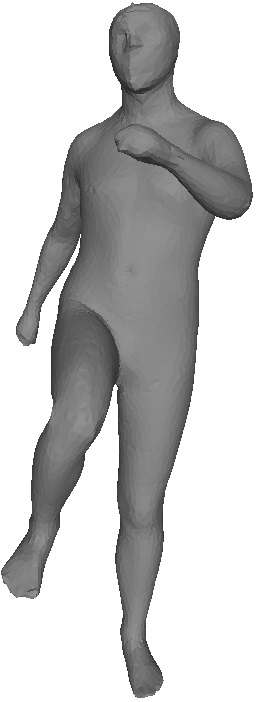} &
\includegraphics[scale=0.12]{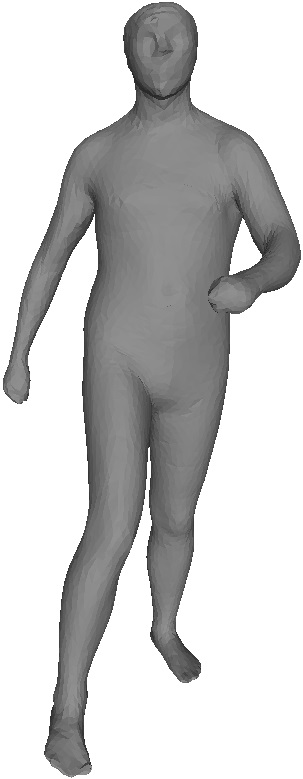} &
\includegraphics[scale=0.12]{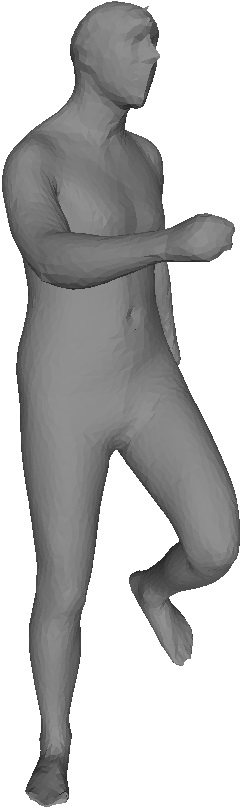} &
\includegraphics[scale=0.12]{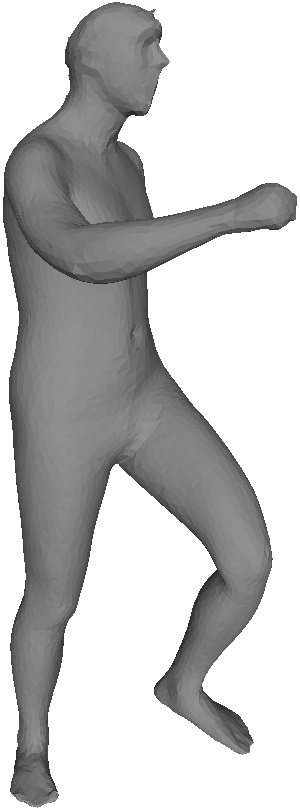} &
\includegraphics[scale=0.12]{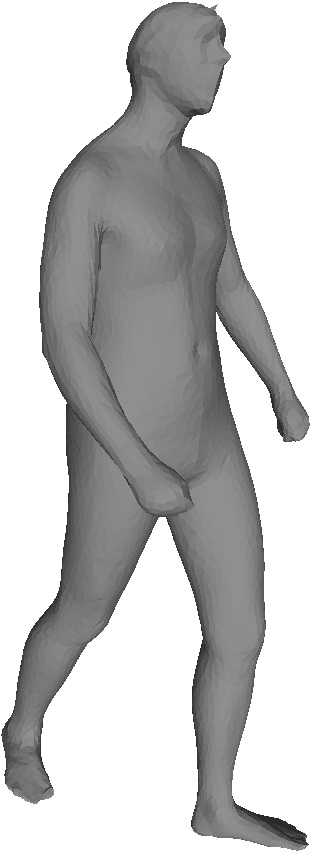} \\
\hline
\includegraphics[scale=0.13]{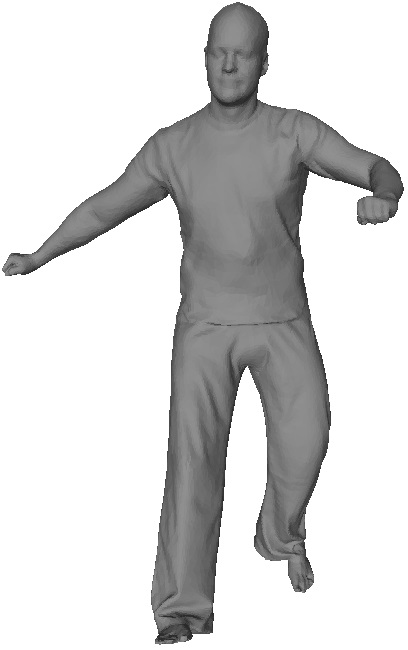} &
\includegraphics[scale=0.13]{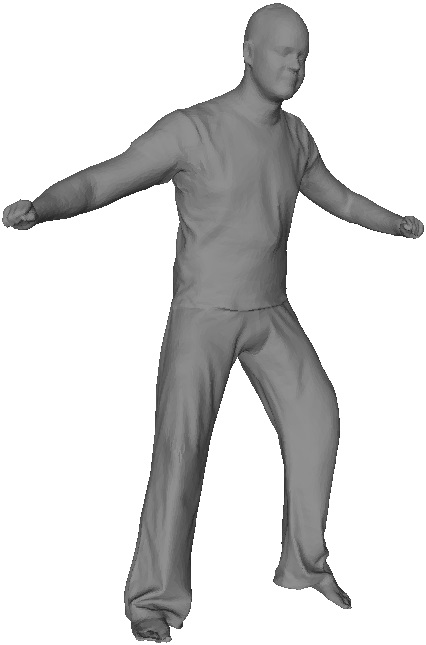} &
\includegraphics[scale=0.13]{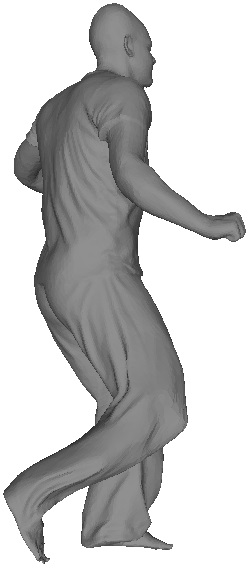} &
\includegraphics[scale=0.13]{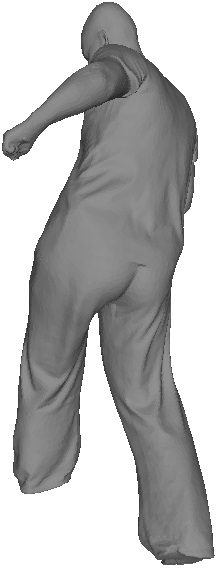} &
\includegraphics[scale=0.13]{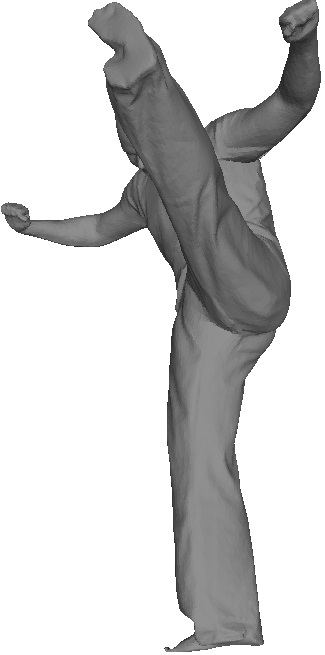} &
\includegraphics[scale=0.13]{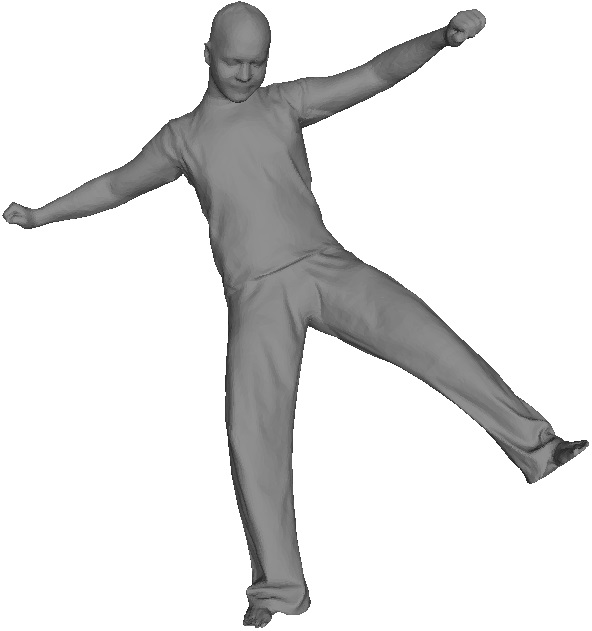} &
\includegraphics[scale=0.13]{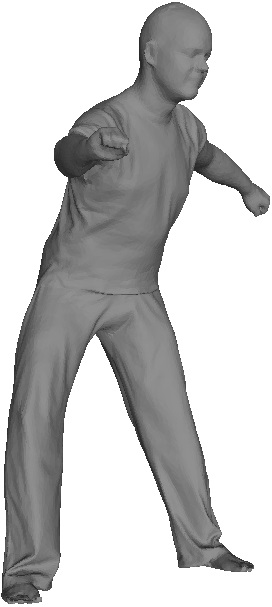} \\
\includegraphics[scale=0.13]{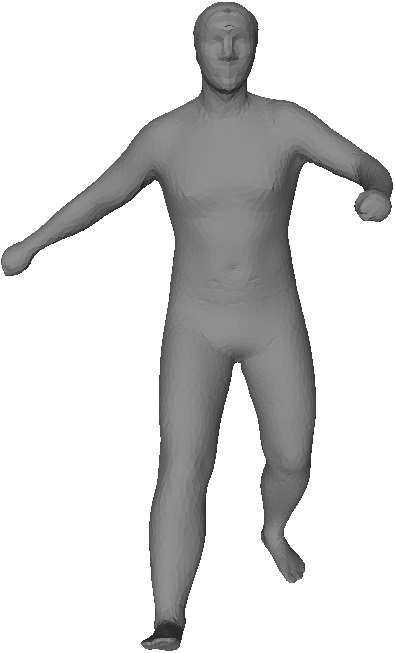} &
\includegraphics[scale=0.13]{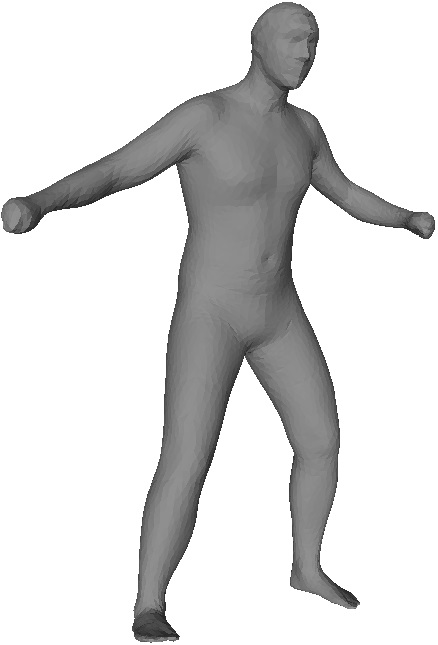} &
\includegraphics[scale=0.13]{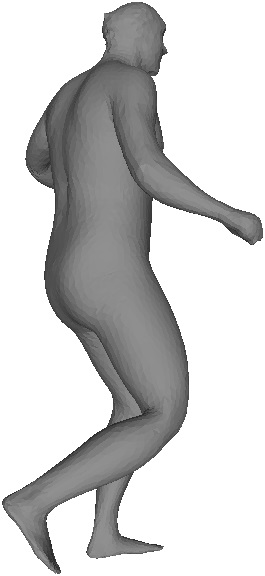} &
\includegraphics[scale=0.13]{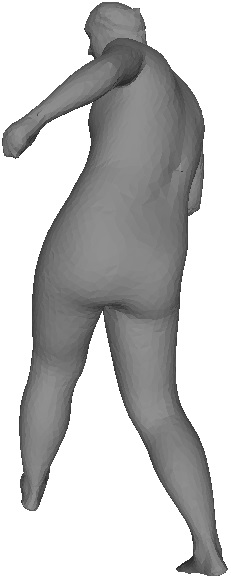} &
\includegraphics[scale=0.13]{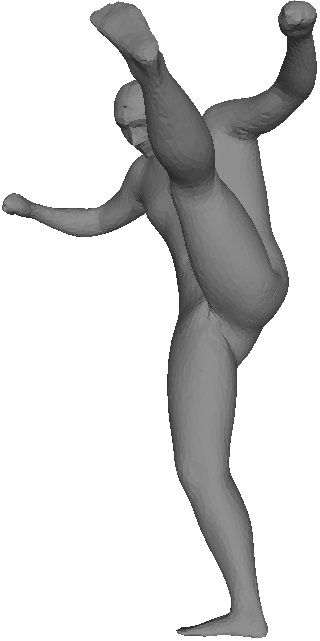} &
\includegraphics[scale=0.13]{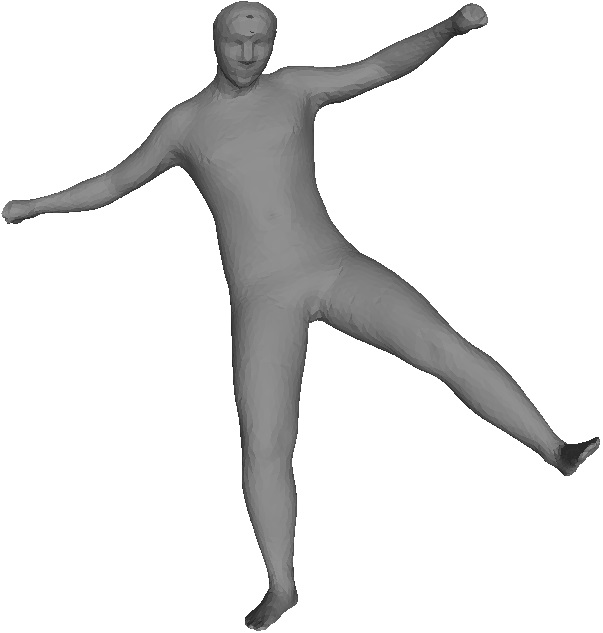} &
\includegraphics[scale=0.13]{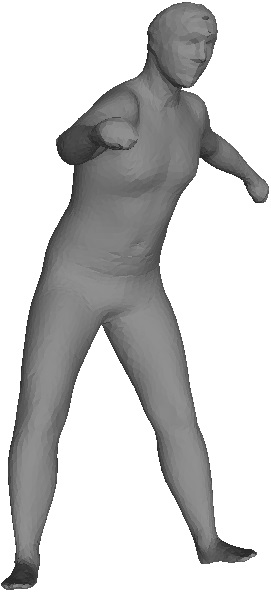} \\
\hline
\includegraphics[scale=0.11]{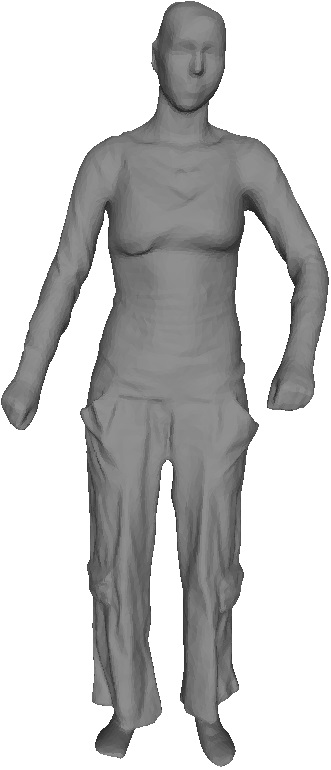} &
\includegraphics[scale=0.11]{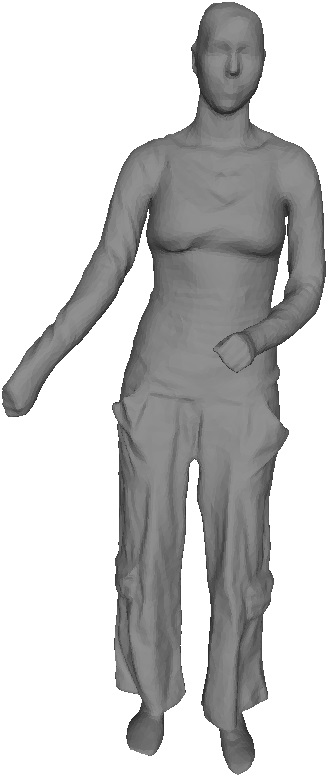} &
\includegraphics[scale=0.11]{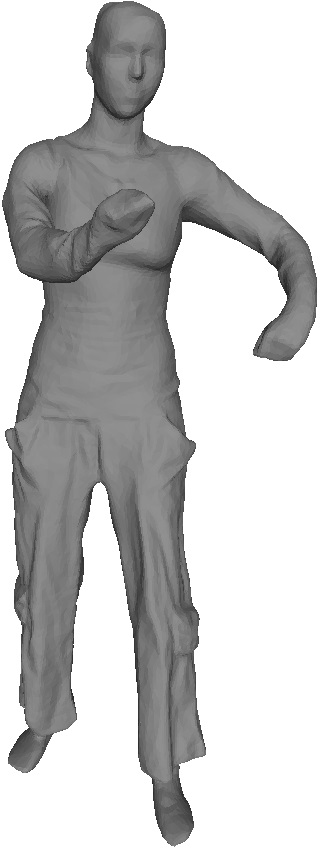} &
\includegraphics[scale=0.11]{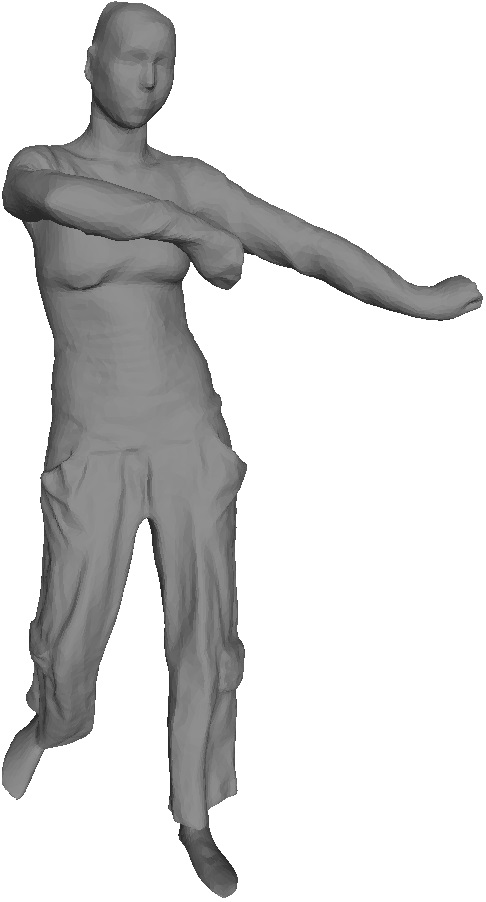} &
\includegraphics[scale=0.11]{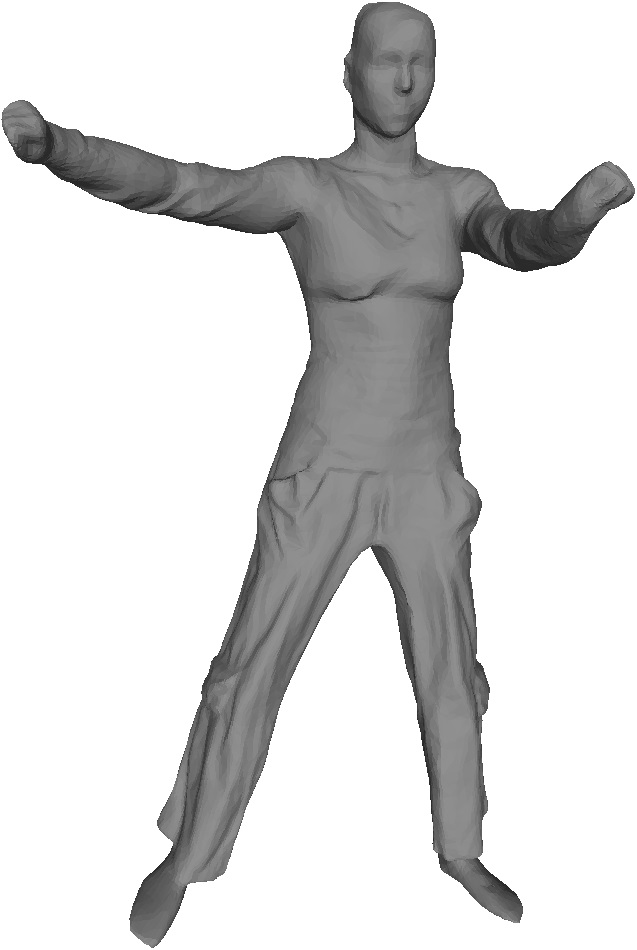} &
\includegraphics[scale=0.11]{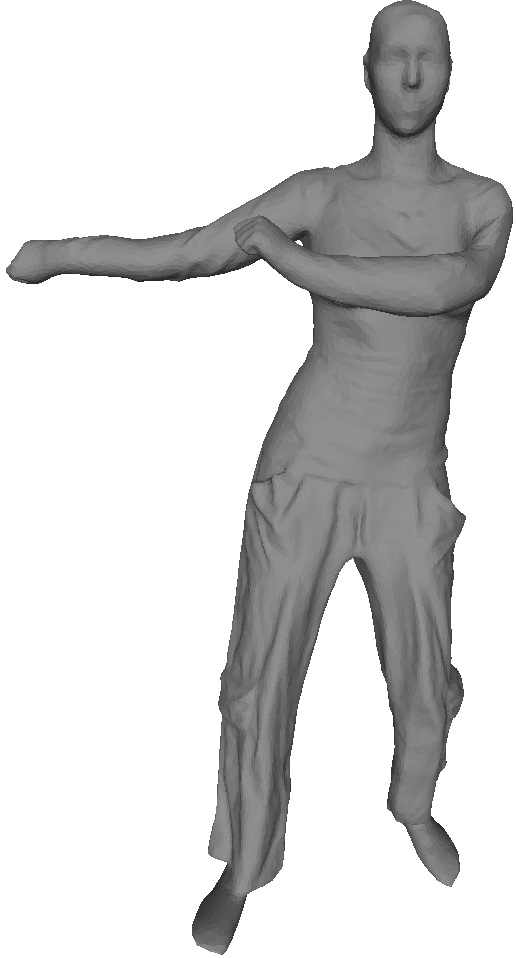} &
\includegraphics[scale=0.11]{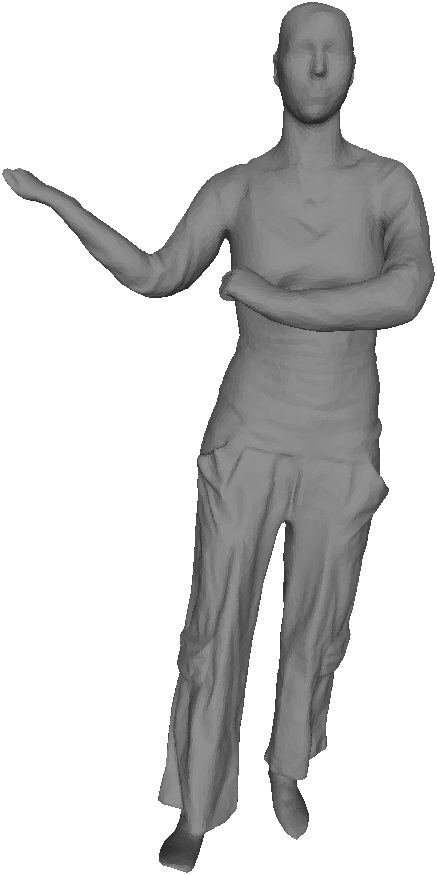} \\
\includegraphics[scale=0.11]{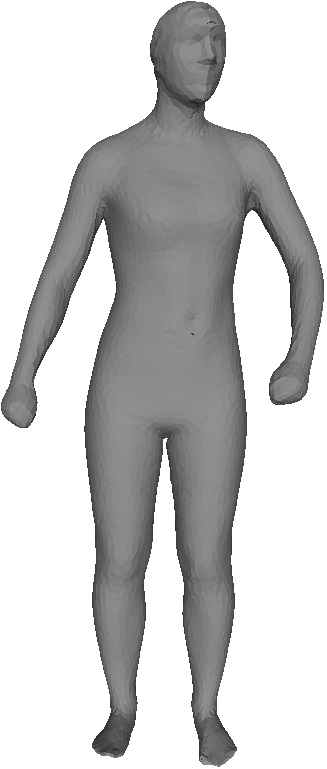} &
\includegraphics[scale=0.11]{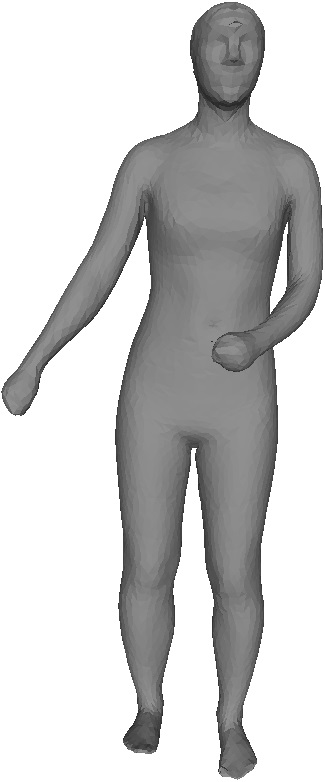} &
\includegraphics[scale=0.11]{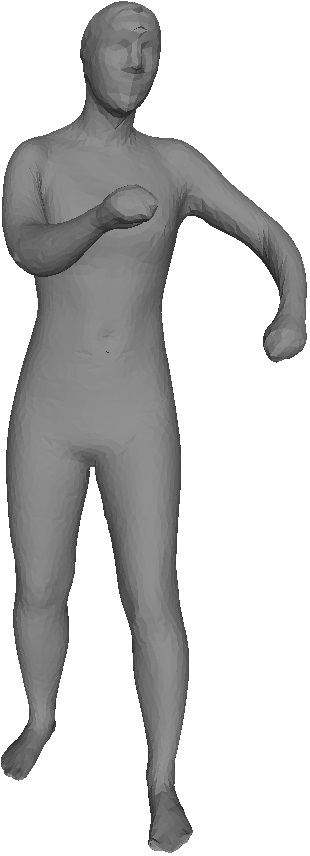} &
\includegraphics[scale=0.11]{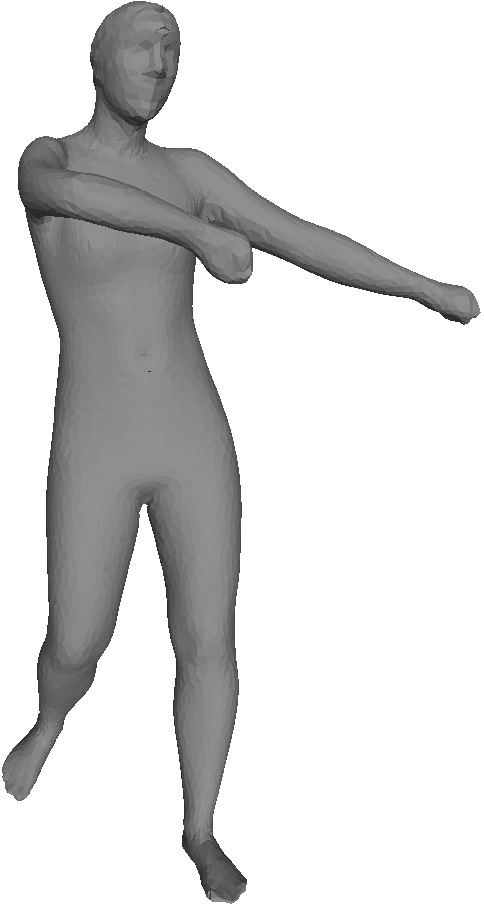} &
\includegraphics[scale=0.11]{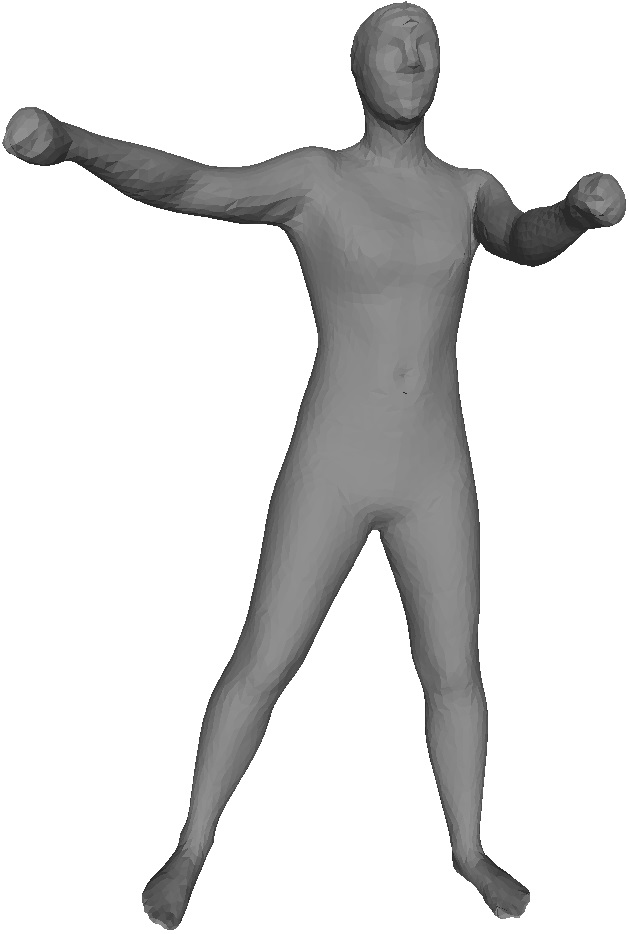} &
\includegraphics[scale=0.11]{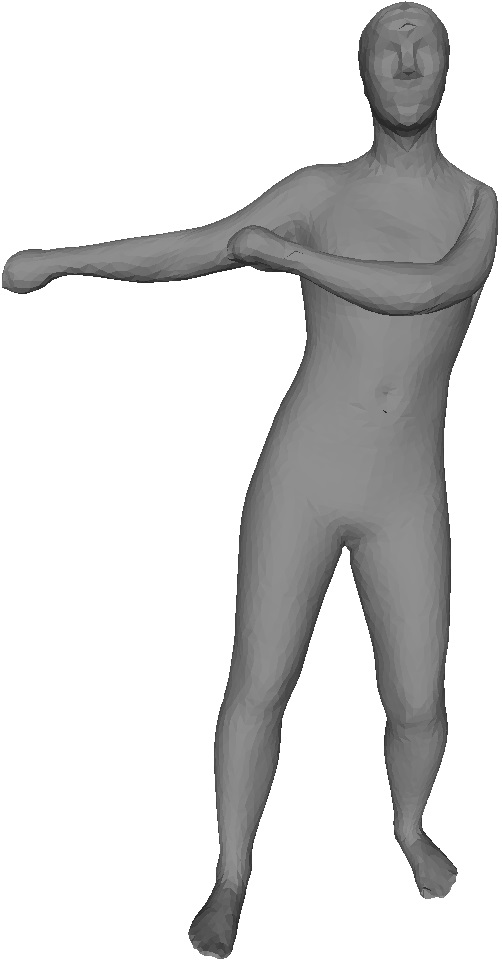} &
\includegraphics[scale=0.11]{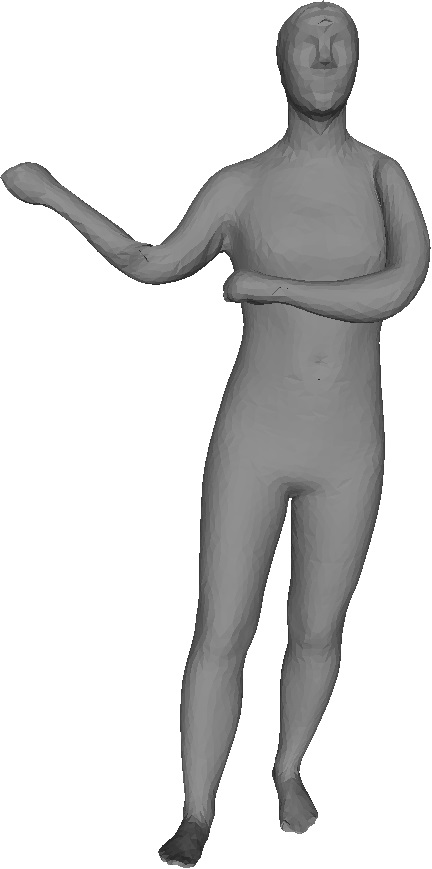} \\
\end{tabular}
\normalsize
\caption{Results of tracking motion sequences acquired using different systems. For each example: top shows the input data and bottom shows our result for seven input frames that are evenly distributed in time.}
\label{fig:results_sequences}
\end{figure}

\paragraph{Acquired motion sequences} We also evaluate our method when fitting the learned statistical model to motion sequences of dressed subjects acquired using different systems. Since there is no ground truth available for this input data, we evaluate the results visually in this case. We fit our model to three input sequences of a male subject acquired while marching~\cite{Vlasic:2008:AMA} (we use a sequence of 57 frames), a male subject acquired performing a kicking motion~\cite{aguiar_etal_track_siggraph_08} (we use a sequence of 39 frames), and a female subject acquired while dancing~\cite{aguiar_etal_track_siggraph_08} (we use a sequence of 49 frames). Figure~\ref{fig:results_sequences} shows the input data and the results of our method for several frames, and results for the full motion sequences can be seen in the supplementary material. Note that in spite of the loose clothing, realistic body shapes are obtained. Furthermore, due to the stable initialization with automatically placed landmarks, the tracking does not fail, even in the case of the fast kicking motion.

\subsection{Limitations}

Finally, we outline some limitations of our method. First, by using a skeleton-based deformation to model posture changes, our method may generate unrealistic bending at joints, especially when the data to fit to is missing or unreliable in this area. This can be observed on the right elbow shown in the rightmost frame in the last row of Figure~\ref{fig:results_sequences}. The reason for such artefacts is that muscle bulging and stretching are not modeled in our shape space.

\begin{figure}[tb]
\centering
\includegraphics[height=3.0cm]{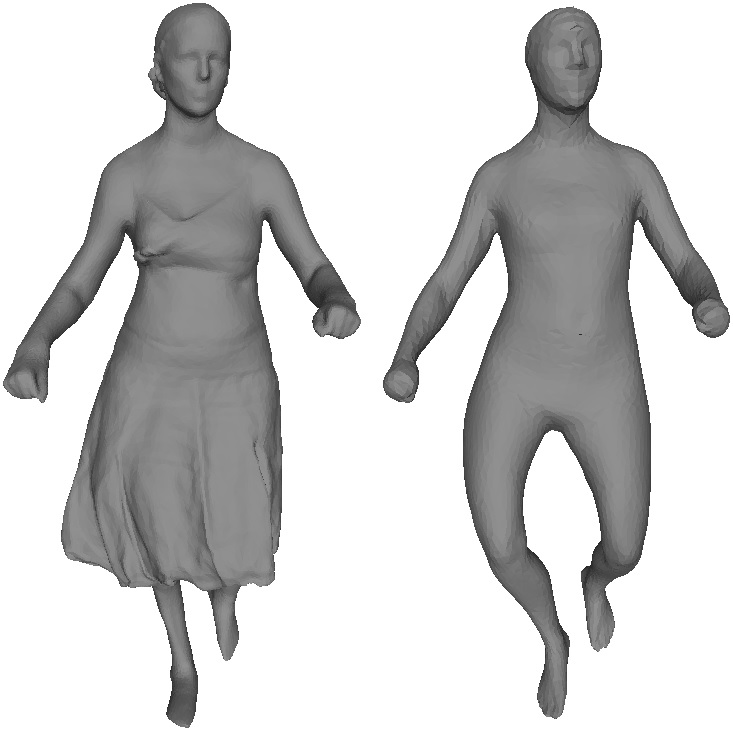}
\caption{Input data and estimated body shape for a frame of a sequence showing a dancing woman wearing a skirt.}
\label{fig:loose_clothing}
\end{figure}

Second, while we have demonstrated that our method can estimate the human body shape and postures for a given input sequence of scans representing a person dressed in regular clothing, our method fails in cases of very loose clothing, such as skirts or dresses. An example where unrealistic body shapes are estimated is shown in Figure~\ref{fig:loose_clothing}, which shows the $17th$ frame of an input sequence of a dancing woman (dataset from de Aguiar et al.~\cite{aguiar_etal_track_siggraph_08}). For this sequence, our method computes a valid output in each step. However, the estimated shape and posture of the upper legs is unrealistic. For more extreme cases like a person wearing a wide dress, where a significant portion of the body is obstructed by loose clothing, we expect the landmark prediction method to fail as the intrinsic geometry of the scan no longer resembles the learned shape space. However, we have not observed this problem in our experiments.

Furthermore, there is currently no guarantee that the estimated body shape is inside the observed clothing, even though this must be the case in reality. However, the clothing term used in Equation~\ref{eq:E_shape} discourages the estimated shape to protrude from the scan, and for our experiments, the estimated shape is almost always entirely contained within the scan. The general limitation of not guaranteeing that the estimated shape is inside the clothing is shared by other methods that use a SCAPE model to find a shape and posture estimate from an input scan or a set of input images.

\section{Conclusion}

We proposed an approach to estimate the body shape and postures of dressed human subjects in motion. Our method, which uses a posture-invariant shape space to model body shape variation combined with a skeleton-based deformation to model posture variation, was shown to have higher fitting accuracy than a popular variant of the commonly used SCAPE model~\cite{anguelov_srinivasan_koller_thrun_rodgers_05_shapecomp,Jain_etal_10_movie_reshape} when fitting to static scans of both dressed and undressed subjects. Furthermore, we showed that our method performs well on motion sequences of dressed subjects.

\section*{Acknowledgments}

We thank Nils Hasler for making the MPI database available, Daniel Vlasic and Christian Theobalt for making their tracking results available, and the volunteers who participated in our scanning experiment. We further thank Gautham Adithya and Monica Vidriales for help in conducting the comparison to the variant of the SCAPE model, and the anonymous reviewers for insightful comments. This work has partially been funded by the Cluster of Excellence \textit{Multimodal Computing and Interaction} within the Excellence Initiative of the German Federal Government.

\end{document}